%% file: acl_latex.tex
\documentclass[11pt]{article}

\usepackage[final]{acl}
\usepackage{times}
\usepackage{latexsym}
\usepackage{multirow}
\usepackage[T1]{fontenc}
\usepackage[utf8]{inputenc}
\usepackage{microtype}

\usepackage{inconsolata}

\usepackage{amsmath}
\usepackage{amssymb}
\usepackage{mathtools}
\usepackage{amsthm}
\usepackage[capitalize,noabbrev]{cleveref}
\usepackage{threeparttable}
\usepackage[table]{xcolor} 
\usepackage{multirow} 
\usepackage{array} 
\usepackage{makecell}
\usepackage{booktabs}
\usepackage{xspace}
\usepackage{tcolorbox}
\tcbuselibrary{skins,breakable}
\usepackage{graphicx}
\usepackage{enumitem}
\usepackage{subcaption}
\usepackage{fontawesome5}

\usepackage{xcolor}
\usepackage{tcolorbox}
\tcbuselibrary{skins,breakable}
\usepackage{tikz}
\usepackage{fontawesome5}
\usepackage{booktabs}

\definecolor{headerblue}{RGB}{41, 128, 185}
\definecolor{metricgreen}{RGB}{39, 174, 96}
\definecolor{alertred}{RGB}{192, 57, 43}
\definecolor{warningyellow}{RGB}{243, 156, 18}
\definecolor{lightgray}{RGB}{245, 245, 245}
\definecolor{darkgray}{RGB}{52, 73, 94}
\definecolor{mutationpurple}{RGB}{142, 68, 173}
\definecolor{crossoverblue}{RGB}{52, 152, 219}

\title{Chain of Mindset: Reasoning with Adaptive Cognitive Modes}

\author{
\textbf{Tianyi Jiang\textsuperscript{1,2*}},
 \textbf{Arctanx An\textsuperscript{1*}},
 \textbf{Hengyi Feng\textsuperscript{1}},
 \textbf{Naixin Zhai\textsuperscript{7}}, \\
 \textbf{Haodong Li\textsuperscript{3}},
 \textbf{Xiaomin Yu\textsuperscript{7}},
 \textbf{Jiahui Liu\textsuperscript{1}},
 \textbf{Hanwen Du\textsuperscript{}},
 \textbf{Shuo Zhang\textsuperscript{7}},\\
 \textbf{Zhi Yang\textsuperscript{4}},
 \textbf{Jie Huang\textsuperscript{4}},
 \textbf{Youhua Li\textsuperscript{6}},
 \textbf{Yongxin Ni\textsuperscript{5}},
 \textbf{Huacan Wang\textsuperscript{7\dag}},
 \textbf{Ronghao Chen\textsuperscript{1,7\dag}}
\\
\\
 \textsuperscript{1}PKU,
 \textsuperscript{2}BJTU,
  \textsuperscript{3}StepFun,
    \textsuperscript{4}SUFE,
    \textsuperscript{5}NUS,
    \textsuperscript{6}CityUHK,
    \textsuperscript{7}QuantaAlpha
    \vspace{2mm}
\\
\small {
    \textbf{\textsuperscript{*}These authors contributed equally to this work.}
}
\\
 \small{
   \textbf{\dag Correspondence:} 
   \href{mailto:wanghuacan17@mails.ucas.ac.cn}{wanghuacan17@mails.ucas.ac.cn},
   \href{mailto:chenronghao@alumni.pku.edu.cn}{chenronghao@alumni.pku.edu.cn}
 }
}

\begin{document}
\maketitle

\vspace{-7mm}

\begin{abstract}
Human problem-solving is never the repetition of a single mindset, by which we mean a distinct mode of cognitive processing. When tackling a specific task, we do not rely on a single mindset; instead, we integrate multiple mindsets within the single solution process. However, existing LLM reasoning methods fall into a common trap: they apply the same fixed mindset across all steps, overlooking that different stages of solving the same problem require fundamentally different mindsets. This single-minded assumption prevents models from reaching the next level of intelligence. To address this limitation, we propose Chain of Mindset (CoM), a training-free agentic framework that enables step-level adaptive mindset orchestration. CoM decomposes reasoning into four functionally heterogeneous mindsets: Spatial, Convergent, Divergent, and Algorithmic. A Meta-Agent dynamically selects the optimal mindset based on the evolving reasoning state, while a bidirectional Context Gate filters cross-module information flow to maintain effectiveness and efficiency. Experiments across six challenging benchmarks spanning mathematics, code generation, scientific QA, and spatial reasoning demonstrate that CoM achieves state-of-the-art performance, outperforming the strongest baseline by 4.96\% and 4.72\% in overall accuracy on Qwen3-VL-32B-Instruct and Gemini-2.0-Flash, while balancing reasoning efficiency. Our code is publicly available at \href{https://github.com/QuantaAlpha/chain-of-mindset}{https://github.com/QuantaAlpha/chain-of-mindset}.
\end{abstract}

\begin{figure}[ht]
\centering
\includegraphics[width=0.50\textwidth]{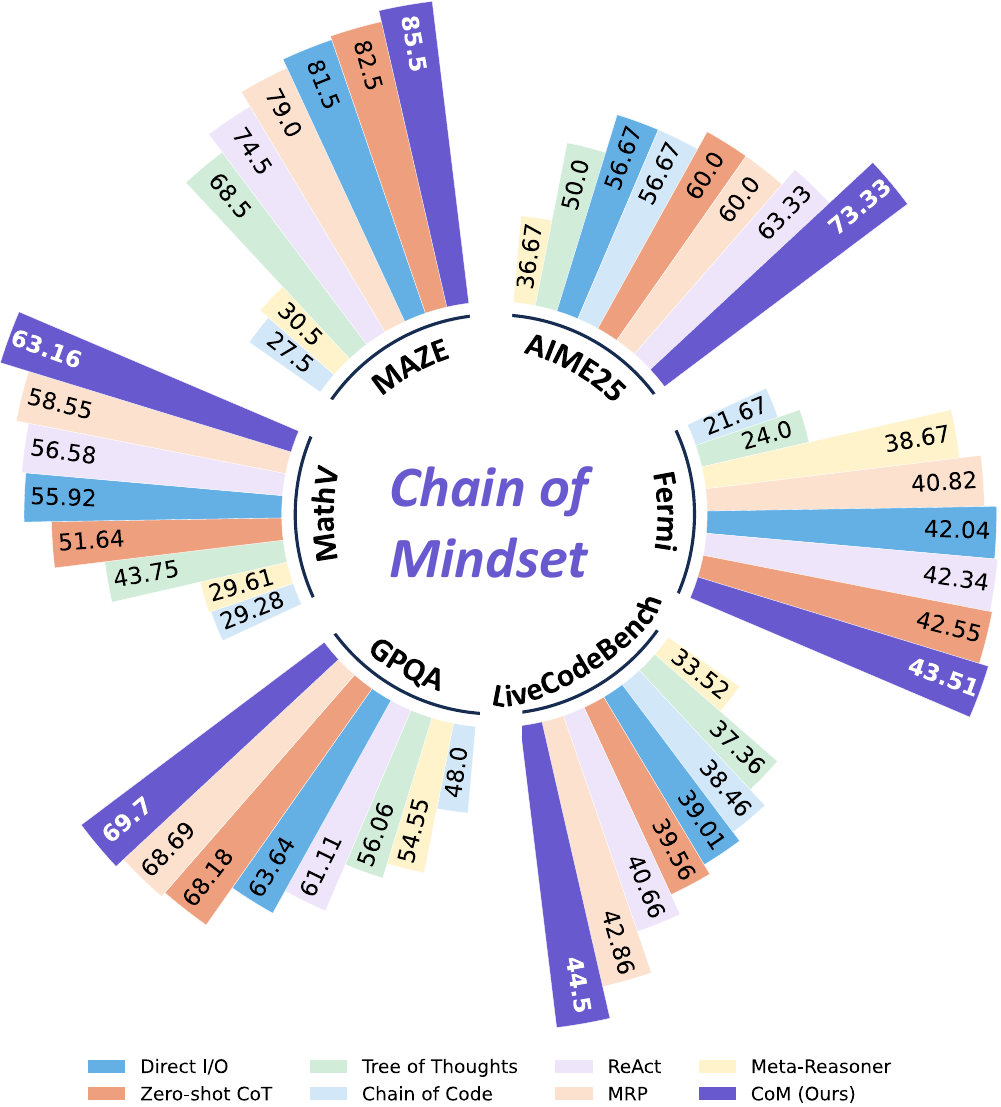}
\caption{Performance comparison on Qwen3-VL-32B-Instruct across six reasoning benchmarks.}
\label{fig:radar_results}
\end{figure}

\newpage

\input{section/1.introduction}

\input{section/2.method}

\input{section/3.experiments}

\input{section/4.conclusion}

\bibliography{reference}

\appendix

\newpage

\appendix
\onecolumn
\label{sec:appendix}
\input{section/x_supp}

\end{document}

%% file: section/1.introduction.tex
\section{Introduction}

The essence of human intelligence lies in the synergy of multiple complementary mindsets. Cognitive science research~\cite{guilford1967nature} has identified distinct cognitive modes that serve fundamentally different functions: Spatial thinking concretizes abstract conditions into intuitive visual representations that facilitate pattern recognition~\cite{newcombe2010picture,newcombe2014thinking}; Convergent thinking distills core insights from complex, multifaceted information through focused logical analysis~\cite{cropley2006praise}; Divergent thinking generates novel possibilities when conventional logic reaches an impasse by exploring unconventional pathways~\cite{runco2012divergent}. This repertoire of cognitive capabilities constitutes the underlying flexibility with which humans handle heterogeneous tasks. Beyond these human cognitive modes, computational systems enable a fourth capability—Algorithmic thinking: precise numerical calculation and formal verification through code execution~\cite{futschek2006algorithmic}, providing computational precision that extends beyond the limits of human mental arithmetic. Yet current intelligent systems, despite their impressive scale and advances in multimodal perception~\cite{lin2025perceive}, lack this repertoire of complementary cognitive capabilities and remain distant from the flexible, multimodal reasoning that characterizes human intelligence.

Crucially, human problem-solving is not merely possessing these mindsets but dynamically orchestrating them within a single reasoning episode~\cite{newell1972human}. When facing a complex task, we do not apply a single mindset uniformly from start to finish; instead, we transition between mindsets as the problem state evolves. For example, solving a geometry proof may begin with spatial reasoning to visualize the configuration, shift to convergent thinking to identify key relationships, then invoke divergent thinking to explore auxiliary constructions, and finally employ algorithmic steps to verify the solution. This step-level adaptive switching (i.e., recognize when each mindset is most effective and transitioning accordingly) is fundamental to human cognitive flexibility~\cite{sali2024learning}. It enables the reasoning trace to remain rigorous when precision is needed and creative when conventional approaches fail.

Previous work~\cite{didolkar2024metacognitive,kargupta2025cognitive} has confirmed that complex reasoning requires diverse mindsets. LLMs indeed exhibit different mindsets during the reasoning process, and prior studies suggest that controlling models through explicit cognitive interventions can effectively improve reasoning performance~\cite{gandhi2025cognitive}. However, a question remains largely unexplored: \textit{Given different contexts and reasoning scenarios, which mindset is most suitable for solving the problem?}

\begin{figure*}[t]
    \centering
    \includegraphics[width=0.7\linewidth]{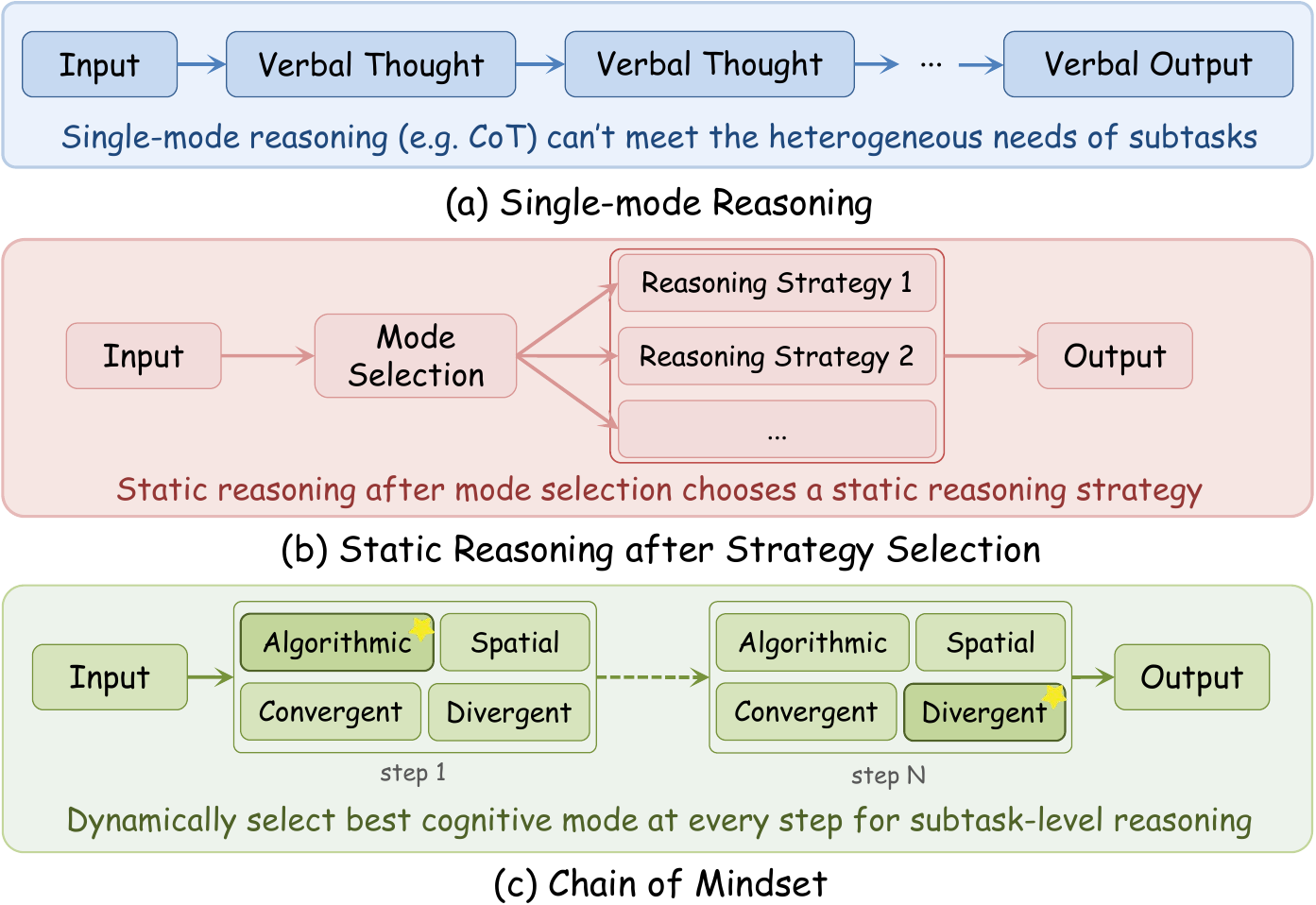}
    \caption{Comparison of reasoning paradigms. (a) Single-mode reasoning applies a single mindset throughout, failing to address heterogeneous sub-task demands. (b)  Static reasoning strategy selection chooses a strategy at task onset but cannot adapt to intermediate states. (c) Chain of Mindset dynamically switches mindsets at subtask boundaries based on the progress of reasoning.}
    \label{fig:paradigm_comparison}
    \vspace{-0.1in}
\end{figure*}

Existing reasoning methods for LLMs fall into two paradigms, both with fundamental limitations, as illustrated in Figure~\ref{fig:paradigm_comparison}. \textbf{Single-mode reasoning} methods~\cite{wei2022chain,chen2022program,li2023chain,yao2023tree} apply a uniform cognitive strategy throughout, struggling when sub-tasks demand heterogeneous capabilities. \textbf{Static reasoning strategy selection} methods~\cite{gao2024mrp,yang2024buffer,aytes2025sketch} choose a reasoning format at task onset but cannot adapt when intermediate results reveal that a different mindset would be more effective. Neither supports dynamic, state-dependent cognitive switching—recognizing when to transition between mindsets based on the progress of reasoning.

To address these limitations, we propose Chain of Mindset (CoM), a training-free agentic reasoning paradigm that implements authentic cognitive chaining. Unlike previous methods that are limited to a single mindset, our framework enables agents to dynamically orchestrate a composite reasoning process with different mindsets. CoM decomposes reasoning into four functionally heterogeneous mindsets—Spatial, Convergent, Divergent, and Algorithmic. We selected these four mindsets because they represent search-style reasoning capabilities that transcend the typical single-mode reasoning of language models.These four mindsets are grounded in foundational cognitive science research as fundamental reasoning paradigms~\cite{guilford1967nature,futschek2006algorithmic,cropley2006praise,newcombe2010picture,runco2012divergent}, each exhibiting distinct behavioral signatures that enable explicit orchestration. When solving any given problem, the agent can adaptively select and dynamically invoke multiple mindsets based on the current state. Furthermore, to prevent cross-boundary information interference caused by frequent mindset switching, we introduce a Context Gate mechanism. Through bidirectional semantic filtering, this mechanism ensures that each thinking module receives only task-relevant context, while the meta-agent receives only highly condensed thought feedback, thereby guaranteeing efficient reasoning. Extensive experiments across six challenging benchmarks demonstrate that CoM consistently outperforms all baselines, as illustrated in Figure~\ref{fig:radar_results}, while maintaining computational efficiency and generalizing across both open-source and closed-source base models without any additional training.



The main contributions are summarized as follows:

\begin{itemize}[noitemsep, topsep=0pt, leftmargin=*]
    \item We propose a new agentic reasoning paradigm. To the best of our knowledge, this is the first training-free method achieve step-level adaptive switching of multiple mindsets within a single inference process.
    \item We formally define four heterogeneous mindsets and propose the Context Gate bidirectional semantic filtering mechanism, which enables the agent to seamlessly switch mindsets while effectively reducing cross-module information interference.
    \item Our experiments on six challenging benchmarks, including mathematics, coding, scientific QA, and spatial reasoning, demonstrate that CoM not only significantly outperforms baseline methods in accuracy but also balances reasoning efficiency with generalization across models and domains without training.
\end{itemize}

%% file: section/2.method.tex
\definecolor{bestcolor}{RGB}{255,235,238}        
\definecolor{secondcolor}{RGB}{227,242,253}      
\definecolor{categorycolor}{RGB}{240,240,240}    

\definecolor{cogdecisioncolor}{RGB}{221, 215, 240}    
\definecolor{algorithmiccolor}{RGB}{214, 238, 250}    
\definecolor{convergentcolor}{RGB}{210, 240, 234}     
\definecolor{spatialcolor}{RGB}{225, 240, 210}        
\definecolor{resultcolor}{RGB}{242, 242, 242}         
\definecolor{insightcolor}{RGB}{240, 215, 235}        
\definecolor{answercolor}{RGB}{220, 245, 220}         
\definecolor{genimgframe}{RGB}{68, 140, 120}          


\section{Method}
\label{sec:method}

In this section, we formally introduce the Chain of Mindset (CoM) framework. We begin by formalizing the mindset switching problem in complex reasoning tasks as a sequential decision-making process in Sec.\ref{sec:problem_formulation}. In Sec.\ref{sec:framework_overview}, we outline the three-layer decoupled architecture of the framework and its design rationale. We then provide detailed definitions of the four heterogeneous mindsets and their cognitive decision mechanism in Sec.\ref{sec:mindset_dispatch}. In Sec.\ref{sec:context_gate}, we demonstrate the necessity and implementation of the Context Gate for inter-module communication from an information-theoretic perspective. Finally, Sec.\ref{sec:case_example} presents an illustrative case study demonstrating the dynamic re-planning capability.

\subsection{Problem Formulation}
\label{sec:problem_formulation}

Consider a geometry proof: one might \textit{spatially} visualize the figure, \textit{divergently} explore auxiliary constructions, \textit{convergently} analyze which approach is promising, and \textit{algorithmically} verify via coordinate calculations. This cognitive flexibility—switching between different mindsets based on intermediate progress—is natural for human experts, yet absent in current LLMs~\cite{kargupta2025cognitive}.

We introduce \textbf{mindset} to formalize distinct cognitive modes. A mindset $m \in \mathcal{M}$ is a specialized reasoning paradigm~\cite{guilford1967nature} characterized by: (1) a distinct cognitive strategy (e.g., parallel exploration vs.\ focused deduction), (2) an isolated context with dedicated prompts, and (3) a structured output. Unlike prior work treating strategies as interchangeable, mindsets are functionally heterogeneous and their deployment requires explicit orchestration.


We define four complementary mindsets $\mathcal{M} = \{m_{\text{spat}}, m_{\text{conv}}, m_{\text{div}}, m_{\text{algo}}\}$, corresponding to spatial imagination, convergent analysis, divergent exploration, and algorithmic computation. Each mindset $m \in \mathcal{M}$ is instantiated through a corresponding call, forming the call set $\mathcal{C} = \{c_{\text{spat}}, c_{\text{conv}}, c_{\text{div}}, c_{\text{algo}}\}$. Given an input problem $q$, the reasoning process unfolds as a trajectory $\mathcal{H} = (c_1, o_1, i_1, c_2, o_2, i_2, \ldots, c_T, o_T, i_T)$, where $c_t \in \mathcal{C}$ denotes the call invoked at step $t$, $o_t$ denotes its output, and $i_t$ denotes its insight. At each step $t$, the agent observes the current state $s_t = (q, \mathcal{H}_{<t})$ and selects the next mindset:
\begin{equation}
    m_t = \pi(s_t) \in \mathcal{M} \cup \{\emptyset\}
\end{equation}
where $\emptyset$ signals termination. The agent then invokes the call $c_t$ corresponding to $m_t$:
\begin{equation}
    (o_t, i_t) = c_t(q, \mathcal{H}_{<t})
\end{equation}
The central insight is that policy $\pi$ conditions on accumulated history $\mathcal{H}_{<t}$: the optimal mindset at step $t$ depends not only on the original problem, but critically on what has been attempted and internalized previously.

To better formularize, we need to address three challenges:
\begin{itemize}[noitemsep, topsep=0pt, leftmargin=*]
    \item \textbf{When to switch}: Judging when the current mindset has exhausted and another would be beneficial.
    \item \textbf{Which mindset to invoke}: Grounding selection in the semantic content of current state rather than relying on surface-level problem type classification.
\item \textbf{How to prevent interference}: Each mindset requires an isolated context, yet must selectively receive relevant information and return distilled results to the Meta-Agent without polluting the main chain.
\end{itemize}

\subsection{Framework Overview}
\label{sec:framework_overview}

\begin{figure*}[t]
    \centering
    \includegraphics[width=\linewidth]{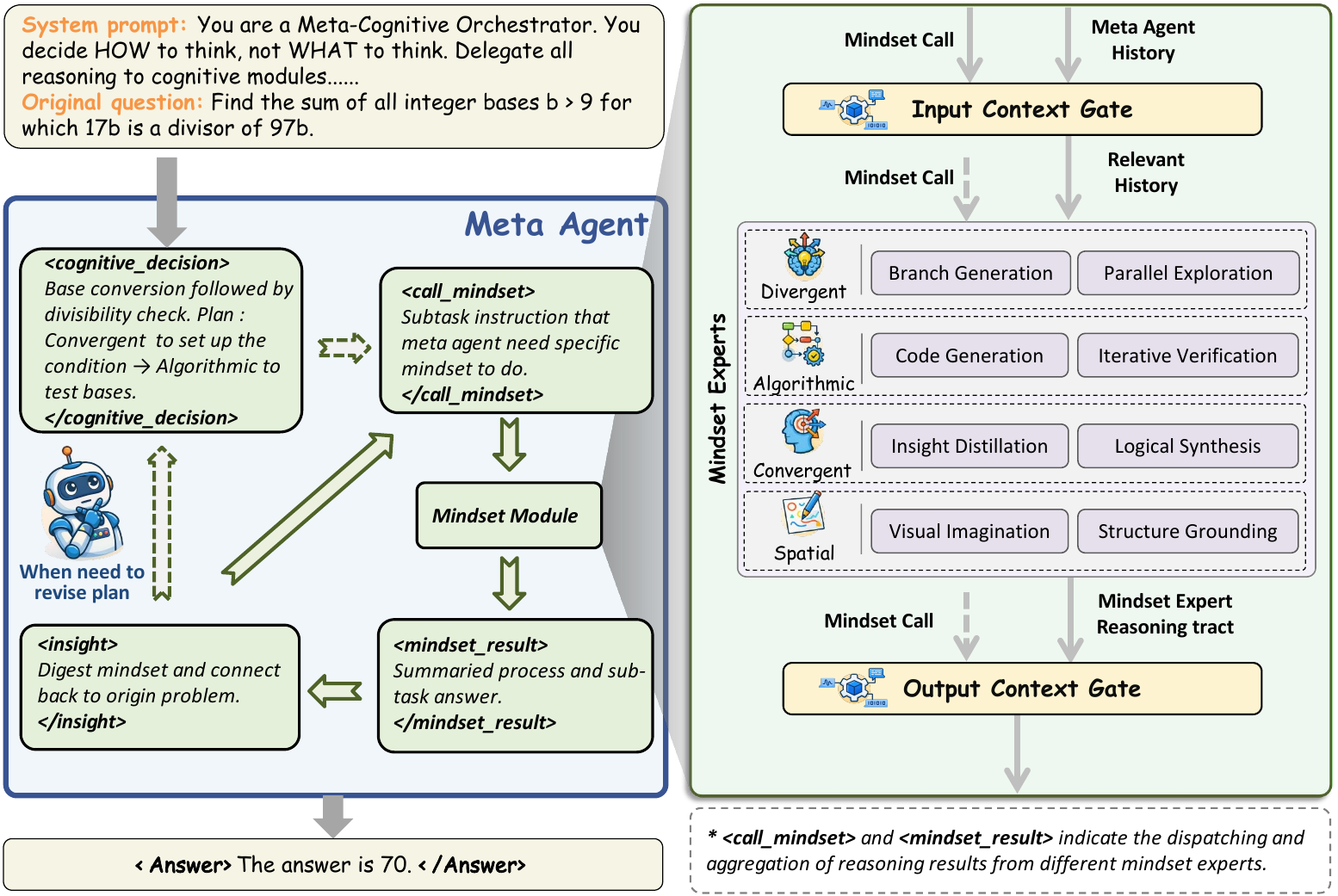}
    \caption{Overview of the Chain of Mindset framework. \textbf{Left}: The Meta-Agent operates as a meta-cognitive orchestrator, iteratively generating cognitive decisions (\texttt{<cognitive\_decision>}), dispatching subtasks to specialized mindsets via call instructions (\texttt{<call\_mindset>}), receiving summarized results (\texttt{<mindset\_result>}), and internalizing key insights (\texttt{<Insight>}) before producing the final answer. The agent may revise its plan when intermediate results warrant replanning. \textbf{Right}: The Mindset Experts comprise four heterogeneous modules—Divergent, Algorithmic, Convergent, and Spatial—each providing distinct cognitive capabilities. The bidirectional Context Gate mediates information flow: the Input Gate filters relevant history for mindset execution, while the Output Gate distills verbose reasoning traces into concise results for the main chain.}
    \label{fig:framework}
    \vspace{-0.1in}
\end{figure*}

To endow the LLM with multi-modal reasoning capabilities while mitigating mutual interference between mindsets, CoM adopts a three-layer decoupled architecture that separates meta-cognitive decision-making from concrete task execution. The framework comprises three core components. First, the Meta-Agent ($\mathcal{A}$) serves as the central controller, orchestrating reasoning by selecting mindsets, generating call instructions, and internalizing intermediate insights. Second, the Mindsets ($\mathcal{M}$) are functionally heterogeneous reasoning modules; each operates within an isolated context, driven by specific system prompts to execute particular sub-tasks. Finally, the Context Gate ($G$) performs bidirectional semantic filtering between the Meta-Agent and Mindsets, mitigating noise from long contexts.

The reasoning process follows an iterative Plan-Call-Internalize loop. Initially, the Meta-Agent generates a cognitive decision $\mathcal{D}$ based on problem characteristics, defining an initial mindset plan. During execution, given state $s_t = (q, \mathcal{H}_{<t})$, the Meta-Agent selects mindset $m_t$ and invokes call $c_t \in \mathcal{C}$ to produce output $o_t$ and insight $i_t$. Upon completion, the Meta-Agent retrieves refined results via the Output Gate and modifies the remaining plan based on newly internalized insight. This mechanism provides flexibility for self-correction within complex reasoning paths.

\begin{figure}[!ht]
\centering
\small
\begin{tcolorbox}[
  colback=white,
  colframe=gray!70,
  boxrule=1pt,
  arc=3pt,
  left=6pt, right=6pt,
  top=5pt, bottom=5pt
]
\textbf{Question \#494:} If the Sun were the head of a body, how long would its arms be?

\textbf{Answer:} \textbf{2,437,190 km}

\tcbline

\colorbox{cogdecisioncolor}{\scriptsize\bfseries\texttt{<cognitive\_decision>}} This is a metaphorical scaling problem---I need to map human body proportions onto the Sun. The key challenge is establishing the correct anatomical ratio. Plan: Spatial (visualize proportions) $\rightarrow$ Convergent (formalize mapping) $\rightarrow$ Algorithmic (compute). \colorbox{cogdecisioncolor}{\scriptsize\bfseries\texttt{</cognitive\_decision>}}

\colorbox{spatialcolor}{\scriptsize\bfseries\texttt{<call\_spatial>}} Generate a human body proportion diagram. I need to see the relative sizes of head and arm to extract the ratio. \colorbox{spatialcolor}{\scriptsize\bfseries\texttt{</call\_spatial>}}

\colorbox{resultcolor}{\scriptsize\bfseries\texttt{<spatial\_result>}} Generated [GEN\_001]. The visualization shows: adult human body $\approx$ 7.5 head-heights tall; arm span $\approx$ body height; single arm (shoulder to fingertip) $\approx$ 3.5 $\times$ head size. \colorbox{resultcolor}{\scriptsize\bfseries\texttt{</spatial\_result>}}

\begin{center}
\begin{tcolorbox}[
  enhanced,
  colback=white,
  colframe=genimgframe,
  boxrule=1pt,
  arc=2pt,
  width=0.80\linewidth,
  left=2pt, right=2pt,
  top=2pt, bottom=2pt,
  title={\footnotesize\ttfamily Generated [GEN\_001]},
  coltitle=genimgframe,
  colbacktitle=white,
  attach boxed title to top right={yshift=-1mm, xshift=-2mm},
  boxed title style={boxrule=0pt, colback=white}
]
\centering
\includegraphics[width=0.8\linewidth]{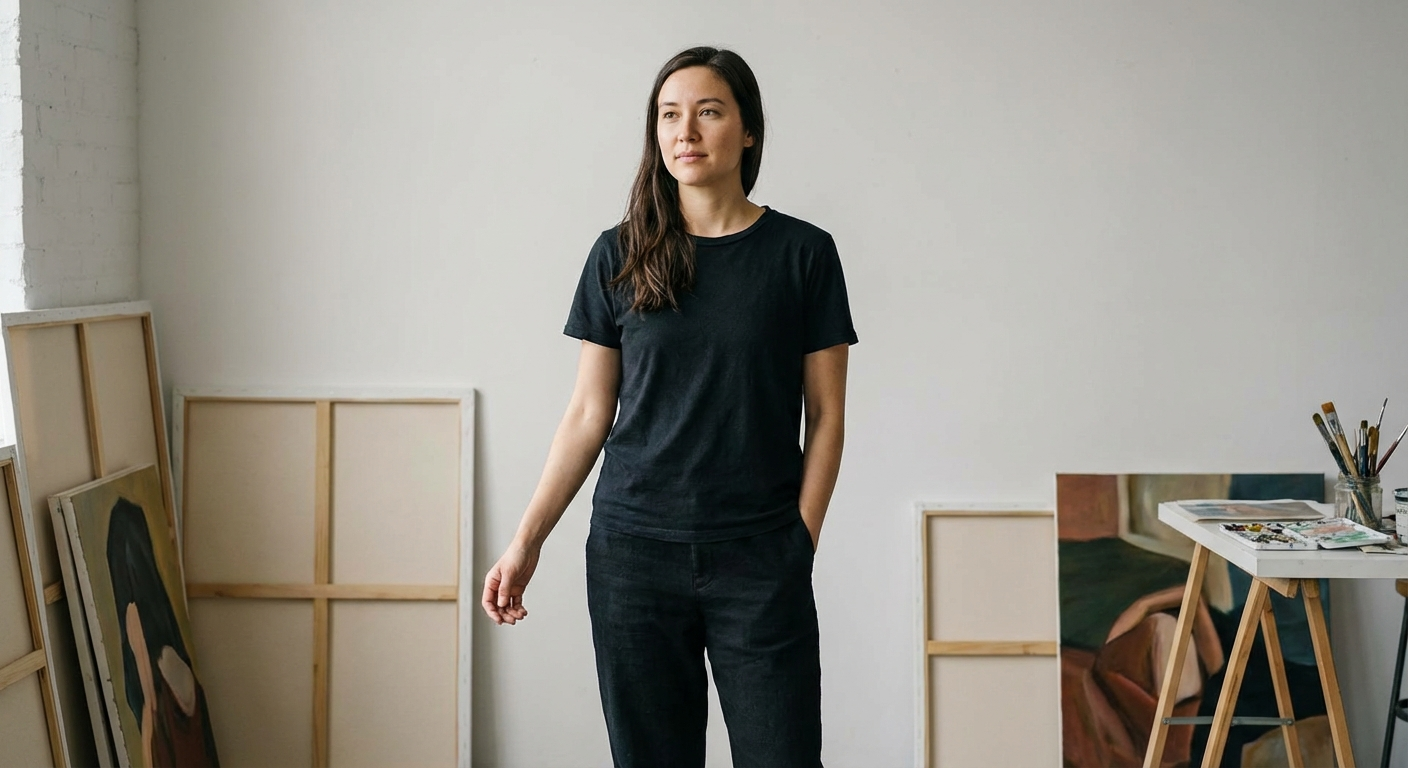}
\end{tcolorbox}
\end{center}

\colorbox{insightcolor}{\scriptsize\bfseries\texttt{<insight>}} From the diagram: arm length $\approx$ 3.5 $\times$ head size. But what does ``head size'' mean here---diameter or radius? The Sun as a ``head'' suggests treating its radius as the characteristic size. \colorbox{insightcolor}{\scriptsize\bfseries\texttt{</insight>}}

\colorbox{convergentcolor}{\scriptsize\bfseries\texttt{<call\_convergent>}} Clarify the mapping: when we say ``the Sun is the head,'' should I use Sun's radius or diameter? \colorbox{convergentcolor}{\scriptsize\bfseries\texttt{</call\_convergent>}}

\colorbox{resultcolor}{\scriptsize\bfseries\texttt{<convergent\_result>}} ``Size'' in anatomical context typically refers to the characteristic dimension---for a spherical head, this is the radius (half-width). Sun's radius $R_{\odot}$ = 696,340 km. \colorbox{resultcolor}{\scriptsize\bfseries\texttt{</convergent\_result>}}

\colorbox{insightcolor}{\scriptsize\bfseries\texttt{<insight>}} Mapping confirmed: Sun radius $\rightarrow$ head size. Arm length = $3.5 \times R_{\odot}$. \colorbox{insightcolor}{\scriptsize\bfseries\texttt{</insight>}}

\colorbox{algorithmiccolor}{\scriptsize\bfseries\texttt{<call\_algorithmic>}} Compute $3.5 \times 696{,}340$. \colorbox{algorithmiccolor}{\scriptsize\bfseries\texttt{</call\_algorithmic>}}

\colorbox{resultcolor}{\scriptsize\bfseries\texttt{<algorithmic\_result>}} $3.5 \times 696{,}340 = 2{,}437{,}190$ km. \colorbox{resultcolor}{\scriptsize\bfseries\texttt{</algorithmic\_result>}}

\colorbox{answercolor}{\scriptsize\bfseries\texttt{<Answer>}} \textbf{2,437,190 km} \colorbox{answercolor}{\scriptsize\bfseries\texttt{</Answer>}}
\end{tcolorbox}
\vspace{-0.1in}
\caption{Fermi problem (\#494) demonstrating the Spatial Mindset. The Spatial Mindset generates an anatomy diagram to visually ground the abstract proportion and extract the head-to-arm ratio ($\approx 3.5\times$). The subsequent Convergent call resolves an ambiguity: ``head size'' maps to the Sun's radius rather than diameter.}
\label{fig:case_example}
\vspace{-0.1in}
\end{figure}

\subsection{Mindset Dispatch}
\label{sec:mindset_dispatch}

Each mindset receives a filtered input tuple from the Input Gate: the call instruction $c$, relevant context $\mathcal{H}_{\text{rel}}$ extracted from reasoning history, and injected images $\mathcal{I}_{\text{inj}}$ when visual information is required. We define four complementary mindsets, each instantiated as a specialized execution module with distinct cognitive strategies.

\textbf{Spatial Mindset} ($m_{\text{spat}}$). This mindset bridges abstract logic and intuitive perception through visual externalization~\cite{lin2024draw,li2025mvot,zhang2025latent}. Given instruction $c$, we supports three generation modes via Nano-Banana-Pro~\cite{google2025nanobanana}: (1) \textit{Text}$\rightarrow$\textit{Image}: pure textual descriptions are transformed into visualizations; (2) \textit{Image+Text}$\rightarrow$\textit{Image}: referenced images $\mathcal{I}_{\text{inj}}$ are edited or augmented based on $c$; (3) \textit{Code}$\rightarrow$\textit{Image}: when the model returns matplotlib code, it is executed in a sandbox to produce figures. Generated artifacts are registered into a global library with unique identifiers (e.g., \texttt{[GEN\_001]}) for reference in subsequent reasoning steps.

\textbf{Convergent Mindset} ($m_{\text{conv}}$). This mindset addresses information overload by constructing a focused reasoning environment~\cite{wei2022chain,pan2023logic}. Given instruction $c$ and filtered context $\mathcal{H}_{\text{rel}}$, it performs a single deep reasoning pass that grounds each step in established facts, explicitly states missing information, and reaches a clear conclusion. The output is a complete logical derivation.

\textbf{Divergent Mindset} ($m_{\text{div}}$). This mindset breaks reasoning deadlocks through structured parallel exploration~\cite{wang2022self,yao2023tree}. Given instruction $c$ and filtered context $\mathcal{H}_{\text{rel}}$, execution proceeds in two phases: (1) \textit{Branch Generation}: produce $k \in [2,5]$ distinct solution branches $\{b_1, \ldots, b_k\}$, where each branch $b_i$ represents a candidate reasoning path with explicit assumptions; (2) \textit{Parallel Exploration}: independently analyze each $b_i$ through a separate LLM call that examines its step-by-step procedure and potential limitations. Crucially, all branch explorations $\{r_1, \ldots, r_k\}$ are returned to the Meta-Agent $\mathcal{A}$ for path selection, preserving deliberation at the metacognitive level.

\textbf{Algorithmic Mindset} ($m_{\text{algo}}$). This mindset addresses limitations of language models in precise calculation through a code-based generate-execute-repair loop~\cite{chen2022program,gao2023pal,li2023chain}. Given instruction $c$ and filtered context $\mathcal{H}_{\text{rel}}$, let $\rho_0$ denote the initially generated Python code. The execution iterates:
\begin{equation}
    (\rho_{i+1}, r_{\text{algo}}) = \begin{cases}
    (\rho_i, \textsc{Exec}(\rho_i)) & \text{if execution succeeds} \\
    (\textsc{Fix}(\rho_i, \epsilon_i), \bot) & \text{if error } \epsilon_i \land i < N_{\max} \\
    (\rho_i, \epsilon_i) & \text{otherwise}
    \end{cases}
\end{equation}
where $\epsilon_i$ denotes the execution error at iteration $i$, $N_{\max} = 2$ bounds repair attempts, and $\bot$ indicates pending status before the next iteration.

All mindsets produce a unified output tuple $(r, \mathcal{I}_{\text{new}})$ passed to the Output Gate, where $r$ contains the reasoning trace or execution log, and $\mathcal{I}_{\text{new}}$ denotes newly generated visual artifacts. The Output Gate distills this verbose output into a concise summary $O_{\text{sum}}$, which the Meta-Agent internalizes as \texttt{<insight>} to integrate into the main reasoning chain.

\subsection{Context Gate}
\label{sec:context_gate}

In modular reasoning systems, information transfer faces a Relevance-Redundancy Trade-off~\cite{liu2024lost}. Directly transmitting the complete history leads to context pollution, while transmitting only instructions results in context starvation. We address this issue from the perspective of Information Density. Let the reasoning history at time step $t$ be $\mathcal{H}_t$ and the call instruction be $c_t$. In the input direction, the effective information density is defined as $\rho_{\text{in}} = |\mathcal{H}_{\text{rel}}| / |\mathcal{H}_t|$, where $\mathcal{H}_{\text{rel}}$ is the context subset relevant to the sub-task. As the reasoning step $t$ increases, $\rho_{\text{in}} \to 0$, implying a linear growth in noise~\cite{li2024long}. In the output direction, the raw output $r$ of a mindset often contains extensive intermediate processes, whereas the main chain requires only key conclusions $O_{\text{sum}}$, leading to an output density $\rho_{\text{out}} \ll 1$.

The design objective of the Context Gate is to increase bidirectional information density ($\rho \to 1$). This mechanism consists of two components, each driven by an independent LLM. The Input Gate ($G_{\text{in}}$) uses the call instruction $c$ as a semantic anchor to extract the minimal sufficient context set $\mathcal{H}_{\text{rel}}$ and relevant images $\mathcal{I}_{\text{inj}}$ from history $\mathcal{H}$:
\begin{equation}
    (\mathcal{H}_{\text{rel}}, \mathcal{I}_{\text{inj}}) = G_{\text{in}}(\mathcal{H}, c, M, \mathcal{I})
\end{equation}
Conversely, the Output Gate ($G_{\text{out}}$) distills the key insight $O_{\text{sum}}$ from the verbose mindset output $r$ based on the expected goal of instruction $c$:
\begin{equation}
    O_{\text{sum}} = G_{\text{out}}(r, c, \mathcal{I}_{\text{new}})
\end{equation}
Through this bidirectional semantic filtering, the Context Gate ensures the efficient execution of mindsets in isolated environments while maintaining the compactness of the main reasoning chain. Complete prompt templates for all components are provided in Appendix~\ref{app:com_prompts}.

\subsection{Illustrative Example}
\label{sec:case_example}

We demonstrate CoM on a Fermi estimation problem (\#494), showcasing how mindset switching enables natural reasoning: Spatial grounds abstract proportions through visualization, Convergent resolves semantic ambiguity, and Algorithmic ensures computational precision.

This illustrates two key capabilities. First, \textit{visual grounding}: CoM externalizes abstract quantities as verifiable images rather than relying on parametric recall. Second, \textit{ambiguity resolution}: the internalization mechanism enables the Meta-Agent to detect underspecified mappings and trigger targeted clarification. Additional cases demonstrating other capabilities (e.g., dynamic re-planning, multimodal input) are provided in Appendix~\ref{app:case_studies}.

%% file: section/3.experiments.tex
\definecolor{bestcolor}{RGB}{255,235,238}        
\definecolor{secondcolor}{RGB}{227,242,253}      
\definecolor{categorycolor}{RGB}{240,240,240}    

\section{Experiments}
\label{sec:experiments}

We evaluate CoM through experiments on diverse reasoning tasks. We first describe the tasks and datasets in Sec.~\ref{sec:tasks}, the baselines in Sec.~\ref{sec:baselines}, and the implementation details in Sec.~\ref{sec:impl}. We then present main results in Sec.~\ref{sec:results}, ablation studies in Sec.~\ref{sec:ablation}, and analysis in Sec.~\ref{sec:analysis}.

\subsection{Tasks and Datasets}
\label{sec:tasks}

We evaluate CoM on six benchmarks spanning four categories: (1)~\textbf{Mathematical Reasoning.} \textit{AIME 2025}: All 30 problems from the 2025 American Invitational Mathematics Examination, covering algebra, geometry, combinatorics, and number theory. \textit{Real-Fermi}~\cite{kalyan2021fermi}: 557 Fermi estimation problems requiring order-of-magnitude reasoning (e.g., ``How much coffee was consumed during EMNLP 2019?''). (2)~\textbf{Code Generation.} \textit{LiveCodeBench}~\cite{jain2024livecodebench}: 182 problems from LeetCode, AtCoder, and CodeForces published between January 1 and May 1, 2025, spanning 45 Easy, 55 Medium, and 82 Hard problems. (3)~\textbf{Science QA.} \textit{GPQA}~\cite{rein2024gpqa}. We select a subset called GPQA-Diamond, containing 198 PhD-level questions in physics, chemistry, and biology, for which non-experts achieve only $\sim$30\% accuracy. (4)~\textbf{Multimodal Reasoning.} \textit{MathV}~\cite{wang2024mathvision}: We select a subset called MathVision-Mini containing 152 questions of the multimodal mathematical benchmark requiring visual diagram understanding before symbolic derivation. \textit{MAZE}: 200 maze navigation problems following the protocol of MVoT~\cite{li2025mvot,zhang2025latent}, generated using maze-dataset~\cite{ivanitskiy2023maze}, where models choose the final position after executing a given action sequence from the starting point on a maze image.

\subsection{Baselines}
\label{sec:baselines}
We compare CoM against four categories of methods: (1) \textbf{Direct Reasoning}: Direct I/O and Zero-shot CoT~\cite{kojima2022large} provide fundamental references for single-pass reasoning without explicit orchestration; (2) \textbf{Structured Reasoning}: Tree of Thoughts~\cite{yao2023tree}, Chain of Code~\cite{li2023chain} represent explicit reasoning organization through predetermined structures but lack dynamic adaptation; (3) \textbf{Agentic Reasoning}: ReAct~\cite{yao2022react} equipped with the same Python interpreter and image generation tools as CoM, enable tool use and iterative refinement yet apply a uniform cognitive strategy throughout; (4) \textbf{Meta-Reasoning}: MRP~\cite{gao2024mrp} selects strategies only at task onset, while Meta-Reasoner~\cite{sui2025meta}, though step-level, modulates execution parameters rather than cognitive modes. Detailed implementation settings for each baseline are provided in Appendix~\ref{app:baselines}.

\begin{table*}[!ht]
\caption{Main results on Qwen3-VL-32B-Instruct. We report pass@1 accuracy (\%) across six benchmarks. Overall is the arithmetic mean across benchmarks. 
\colorbox{bestcolor}{\textbf{Bold}}~indicates best; \colorbox{secondcolor}{\underline{underline}}~indicates second best.}
\label{tab:main-qwen}
\begin{center}
\begin{footnotesize}
\setlength{\tabcolsep}{3.5pt}
\resizebox{0.95\textwidth}{!}{%
\begin{tabular}{c|ccccccccc|c}
\toprule
\multirow{2}{*}{\textbf{Method}} & \multicolumn{2}{c}{\textbf{Math}} & \multicolumn{4}{c}{\textbf{LiveCodeBench}} & \textbf{Science} & \multicolumn{2}{c|}{\textbf{Multimodal}} & \multirow{2}{*}{\textbf{Overall}} \\
\cmidrule(lr){2-3} \cmidrule(lr){4-7} \cmidrule(lr){8-8} \cmidrule(lr){9-10}
& AIME25 & Fermi & Easy & Medium & Hard & All & GPQA & MathV & MAZE & \\
\midrule
Direct I/O & 56.67 & 42.04 & \cellcolor{secondcolor}\underline{95.56} & 36.36 & 9.76 & 39.01 & 63.64 & 55.92 & 81.50 & 56.46 \\
Zero-shot CoT & 60.00 & \cellcolor{secondcolor}\underline{42.55} & \cellcolor{bestcolor}\textbf{97.78} & 36.36 & 9.76 & 39.56 & 68.18 & 51.64 & \cellcolor{secondcolor}\underline{82.50} & 57.41 \\
Tree of Thoughts & 50.00 & 24.00 & 86.67 & 34.55 & 12.20 & 37.36 & 56.06 & 43.75 & 68.50 & 46.61 \\
Chain of Code & 56.67 & 21.67 & 88.89 & 38.18 & 10.98 & 38.46 & 48.00 & 29.28 & 27.50 & 36.93 \\
ReAct & \cellcolor{secondcolor}\underline{63.33} & 42.34 & 88.89 & \cellcolor{secondcolor}\underline{40.00} & 14.63 & 40.66 & 61.11 & 56.58 & 74.50 & 56.42 \\
MRP & 60.00 & 40.82 & \cellcolor{secondcolor}\underline{95.56} & 36.36 & \cellcolor{bestcolor}\textbf{18.29} & \cellcolor{secondcolor}\underline{42.86} & \cellcolor{secondcolor}\underline{68.69} & \cellcolor{secondcolor}\underline{58.55} & 79.00 & \cellcolor{secondcolor}\underline{58.32} \\
Meta-Reasoner & 36.67 & 38.67 & 80.00 & 34.55 & 7.32 & 33.52 & 54.55 & 29.61 & 30.50 & 37.25 \\
\textbf{CoM (Ours)} & \cellcolor{bestcolor}\textbf{73.33} & \cellcolor{bestcolor}\textbf{43.51} & 93.33 & \cellcolor{bestcolor}\textbf{45.45} & \cellcolor{secondcolor}\underline{17.07} & \cellcolor{bestcolor}\textbf{44.50} & \cellcolor{bestcolor}\textbf{69.70} & \cellcolor{bestcolor}\textbf{63.16} & \cellcolor{bestcolor}\textbf{85.50} & \cellcolor{bestcolor}\textbf{63.28} \\
\bottomrule
\end{tabular}%
}
\end{footnotesize}
\end{center}
\end{table*}

\begin{table*}[!ht]
\caption{Main results on Gemini-2.0-Flash. We report pass@1 accuracy (\%) across six benchmarks. Overall is the arithmetic mean across benchmarks. \colorbox{bestcolor}{\textbf{Bold}}~indicates best; \colorbox{secondcolor}{\underline{underline}}~indicates second best.}
\label{tab:main-gemini}
\begin{center}
\begin{footnotesize}
\setlength{\tabcolsep}{3pt}
\resizebox{0.95\textwidth}{!}{%
\begin{tabular}{c|ccccccccc|c}
\toprule
\multirow{2}{*}{\textbf{Method}} & \multicolumn{2}{c}{\textbf{Math}} & \multicolumn{4}{c}{\textbf{LiveCodeBench}} & \textbf{Science} & \multicolumn{2}{c|}{\textbf{Multimodal}} & \multirow{2}{*}{\textbf{Overall}} \\
\cmidrule(lr){2-3} \cmidrule(lr){4-7} \cmidrule(lr){8-8} \cmidrule(lr){9-10}
& AIME25 & Fermi & Easy & Medium & Hard & All & GPQA & MathV & MAZE & \\
\midrule
Direct I/O & 26.67 & 38.60 & 88.89 & 18.18 & 8.54 & 31.32 & 62.63 & 48.03 & \cellcolor{secondcolor}\underline{76.50} & 47.29 \\
Zero-shot CoT & 23.33 & \cellcolor{secondcolor}\underline{40.92} & 91.11 & 21.82 & 6.10 & 31.87 & 64.14 & 48.36 & 76.00 & 47.44 \\
Tree of Thoughts & 23.33 & 21.47 & 60.00 & 5.45 & 7.31 & 19.78 & 46.46 & 39.14 & 69.50 & 36.61 \\
Chain of Code & \cellcolor{secondcolor}\underline{30.00} & 39.80 & 86.67 & 16.36 & 6.10 & 29.12 & 37.40 & 22.00 & 25.50 & 30.64 \\
ReAct & 23.33 & 37.91 & \cellcolor{secondcolor}\underline{88.89} & \cellcolor{bestcolor}\textbf{40.00} & \cellcolor{bestcolor}\textbf{14.63} & \cellcolor{bestcolor}\textbf{40.66} & 61.62 & 47.37 & 71.50 & 47.07 \\
MRP & 26.67 & 36.09 & \cellcolor{bestcolor}\textbf{95.56} & 20.00 & 6.10 & 32.40 & \cellcolor{secondcolor}\underline{65.15} & \cellcolor{secondcolor}\underline{49.34} & \cellcolor{secondcolor}\underline{76.50} & \cellcolor{secondcolor}\underline{47.69} \\
Meta-Reasoner & 26.67 & 25.31 & 75.56 & 20.00 & 4.88 & 26.92 & 53.54 & 21.05 & 30.50 & 30.67 \\
\textbf{CoM (Ours)} & \cellcolor{bestcolor}\textbf{33.33} & \cellcolor{bestcolor}\textbf{43.05} & \cellcolor{secondcolor}\underline{88.89} & \cellcolor{secondcolor}\underline{36.36} & \cellcolor{secondcolor}\underline{9.75} & \cellcolor{secondcolor}\underline{37.36} & \cellcolor{bestcolor}\textbf{65.70} & \cellcolor{bestcolor}\textbf{51.00} & \cellcolor{bestcolor}\textbf{84.00} &  \cellcolor{bestcolor}\textbf{52.41} \\ 
\bottomrule
\end{tabular}%
}
\end{footnotesize}
\end{center}
\vskip -0.1in
\end{table*}

\begin{table*}[t]
\caption{Ablation study on Qwen3-VL-32B-Instruct. Each row removes one component from the full CoM. We report pass@1 accuracy (\%); Overall is the arithmetic mean. Superscripts {\scriptsize\textcolor{red}{$\downarrow$}/\textcolor{green!60!black}{$\uparrow$}} indicate change relative to full CoM. \colorbox{bestcolor}{\textbf{Bold}}~= best; \colorbox{secondcolor}{\underline{underline}}~= second best.}
\label{tab:ablation}
\vskip 0.1in
\begin{center}
\begin{small}
\setlength{\tabcolsep}{3pt}
\resizebox{\textwidth}{!}{%
\begin{tabular}{c|ccccccccc|c}
\toprule
\multirow{2}{*}{\textbf{Variant}} & \multicolumn{2}{c}{\textbf{Math}} & \multicolumn{4}{c}{\textbf{LiveCodeBench}} & \textbf{Science} & \multicolumn{2}{c|}{\textbf{Multimodal}} & \multirow{2}{*}{\textbf{Overall}} \\
\cmidrule(lr){2-3} \cmidrule(lr){4-7} \cmidrule(lr){8-8} \cmidrule(lr){9-10}
& AIME25 & Fermi & Easy & Medium & Hard & All & GPQA & MathV & MAZE & \\
\midrule
\textbf{CoM (Full)} & \cellcolor{bestcolor}\textbf{73.33} & 43.51 & \cellcolor{bestcolor}\textbf{93.33} & \cellcolor{secondcolor}\underline{45.45} & \cellcolor{bestcolor}\textbf{17.07} & \cellcolor{bestcolor}\textbf{44.50} & \cellcolor{bestcolor}\textbf{69.70} & \cellcolor{bestcolor}\textbf{63.16} & \cellcolor{bestcolor}\textbf{85.50} & \cellcolor{bestcolor}\textbf{63.28} \\
\midrule
w/o Divergent & 56.67\textsuperscript{\tiny\textcolor{red}{$\downarrow$16.66}} & 44.69\textsuperscript{\tiny\textcolor{green!60!black}{$\uparrow$1.18}} & 88.89\textsuperscript{\tiny\textcolor{red}{$\downarrow$4.44}} & 36.36\textsuperscript{\tiny\textcolor{red}{$\downarrow$9.09}} & 13.41\textsuperscript{\tiny\textcolor{red}{$\downarrow$3.66}} & 39.01\textsuperscript{\tiny\textcolor{red}{$\downarrow$5.49}} & 65.05\textsuperscript{\tiny\textcolor{red}{$\downarrow$4.65}} & \cellcolor{secondcolor}\underline{62.17}\textsuperscript{\tiny\textcolor{red}{$\downarrow$0.99}} & 81.00\textsuperscript{\tiny\textcolor{red}{$\downarrow$4.50}} & 58.10\textsuperscript{\tiny\textcolor{red}{$\downarrow$5.18}} \\
w/o Convergent & 60.00\textsuperscript{\tiny\textcolor{red}{$\downarrow$13.33}} & \cellcolor{bestcolor}\textbf{45.32}\textsuperscript{\tiny\textcolor{green!60!black}{$\uparrow$1.81}} & \cellcolor{secondcolor}\underline{91.11}\textsuperscript{\tiny\textcolor{red}{$\downarrow$2.22}} & \cellcolor{secondcolor}\underline{45.45}\textsuperscript{\tiny\textcolor{gray}{$-$0.00}} & 13.41\textsuperscript{\tiny\textcolor{red}{$\downarrow$3.66}} & \cellcolor{secondcolor}\underline{42.31}\textsuperscript{\tiny\textcolor{red}{$\downarrow$2.19}} & \cellcolor{secondcolor}\underline{65.15}\textsuperscript{\tiny\textcolor{red}{$\downarrow$4.55}} & 60.86\textsuperscript{\tiny\textcolor{red}{$\downarrow$2.30}} & 83.50\textsuperscript{\tiny\textcolor{red}{$\downarrow$2.00}} & 59.52\textsuperscript{\tiny\textcolor{red}{$\downarrow$3.76}} \\
w/o Algorithmic & \cellcolor{secondcolor}\underline{70.00}\textsuperscript{\tiny\textcolor{red}{$\downarrow$3.33}} & 43.39\textsuperscript{\tiny\textcolor{red}{$\downarrow$0.12}} & 88.89\textsuperscript{\tiny\textcolor{red}{$\downarrow$4.44}} & \cellcolor{bestcolor}\textbf{47.27}\textsuperscript{\tiny\textcolor{green!60!black}{$\uparrow$1.82}} & 13.41\textsuperscript{\tiny\textcolor{red}{$\downarrow$3.66}} & \cellcolor{secondcolor}\underline{42.31}\textsuperscript{\tiny\textcolor{red}{$\downarrow$2.19}} & 64.65\textsuperscript{\tiny\textcolor{red}{$\downarrow$5.05}} & 60.20\textsuperscript{\tiny\textcolor{red}{$\downarrow$2.96}} & \cellcolor{secondcolor}\underline{84.00}\textsuperscript{\tiny\textcolor{red}{$\downarrow$1.50}} & \cellcolor{secondcolor}\underline{60.76}\textsuperscript{\tiny\textcolor{red}{$\downarrow$2.52}} \\
w/o Spatial & \cellcolor{secondcolor}\underline{70.00}\textsuperscript{\tiny\textcolor{red}{$\downarrow$3.33}} & 41.98\textsuperscript{\tiny\textcolor{red}{$\downarrow$1.53}} & 84.44\textsuperscript{\tiny\textcolor{red}{$\downarrow$8.89}} & 38.18\textsuperscript{\tiny\textcolor{red}{$\downarrow$7.27}} & \cellcolor{secondcolor}\underline{15.85}\textsuperscript{\tiny\textcolor{red}{$\downarrow$1.22}} & 39.56\textsuperscript{\tiny\textcolor{red}{$\downarrow$4.94}} & 63.64\textsuperscript{\tiny\textcolor{red}{$\downarrow$6.06}} & 53.29\textsuperscript{\tiny\textcolor{red}{$\downarrow$9.87}} & 81.00\textsuperscript{\tiny\textcolor{red}{$\downarrow$4.50}} & 58.25\textsuperscript{\tiny\textcolor{red}{$\downarrow$5.03}} \\
w/o Context Gate & 53.33\textsuperscript{\tiny\textcolor{red}{$\downarrow$20.00}} & \cellcolor{secondcolor}\underline{44.88}\textsuperscript{\tiny\textcolor{green!60!black}{$\uparrow$1.37}} & 80.00\textsuperscript{\tiny\textcolor{red}{$\downarrow$13.33}} & 36.36\textsuperscript{\tiny\textcolor{red}{$\downarrow$9.09}} & \cellcolor{bestcolor}\textbf{17.07}\textsuperscript{\tiny\textcolor{gray}{$-$0.00}} & 38.46\textsuperscript{\tiny\textcolor{red}{$\downarrow$6.04}} & 64.14\textsuperscript{\tiny\textcolor{red}{$\downarrow$5.56}} & 54.93\textsuperscript{\tiny\textcolor{red}{$\downarrow$8.23}} & 74.50\textsuperscript{\tiny\textcolor{red}{$\downarrow$11.00}} & 55.04\textsuperscript{\tiny\textcolor{red}{$\downarrow$8.24}} \\
\bottomrule
\end{tabular}%
}
\end{small}
\end{center}
\vskip -0.1in
\end{table*}

\subsection{Implementation Details}
\label{sec:impl}

We evaluate all methods using two base models. \textbf{Qwen3-VL-32B-Instruct}~\cite{Qwen3-VL} is the state-of-the-art open-source vision-language model, which we deploy locally on 8$\times$NVIDIA A100-80GB GPUs. \textbf{Gemini-2.0-Flash}~\cite{comanici2025gemini} is Google's high-performance closed-source non-reasoning multimodal model, accessed via OpenRouter's Google Vertex API. For generations, both models use temperature 0.7 and top\_p 0.95, with max\_tokens set to 32768 for Qwen3-VL-32B-Instruct and 8192 for Gemini-2.0-Flash. The Spatial mode in CoM additionally employs Nano-Banana-Pro~\cite{google2025nanobanana} for image generation. The Algorithmic mode executes Python code in a sandboxed environment with a 30-second timeout; all mindsets share the same base model for fair comparison. We report pass@1 accuracy across all experiments.

\subsection{Main Results}
\label{sec:results}

As shown in \cref{tab:main-qwen,tab:main-gemini}, CoM achieves the highest overall accuracy across both base models: 63.28\% on Qwen3-VL-32B-Instruct and 52.41\% on Gemini-2.0-Flash, outperforming the strongest baseline MRP by 4.96\% and 4.72\%, respectively. Among direct reasoning methods, Zero-shot CoT provides consistent gains over Direct I/O, while ToT and CoC show task-specific strengths. Meta-reasoning approaches (MRP, Meta-Reasoner) outperform direct methods, yet CoM surpasses them across most benchmarks.

The performance gains are most pronounced on tasks requiring flexible mindset adaptation. With Qwen3-VL-32B-Instruct, CoM exceeds the second-best method on AIME25 by 10.00\%, demonstrating the value of multi-path exploration via Divergent mindset. On MAZE spatial reasoning, CoM outperforms MRP on both base models by 6.00\% and 7.50\%, respectively. CoM also maintains strong code generation performance on LiveCodeBench, where Algorithmic mindset enables precise computation. 


\subsection{Ablation Study}
\label{sec:ablation}

\cref{tab:ablation} presents ablation results by systematically removing each component from full CoM. The Context Gate proves most critical: its removal causes the largest overall drop of 8.24\%, confirming that adaptive information filtering between meta-agent and mindset experts is essential for effective coordination. Among the four mindsets, Divergent contributes most to mathematical reasoning, with AIME25 accuracy dropping 16.66\% upon removal, while Spatial shows the largest impact on visual tasks, reducing MathVision by 9.87\% and MAZE by 4.50\%. Algorithmic primarily benefits code generation, with LiveCodeBench All dropping 2.19\% when removed.

On Fermi estimation, removing Divergent (+1.18\%), Convergent (+1.81\%), or Context Gate (+1.37\%) all yield slight improvements, while only Algorithmic and Spatial mindsets remain essential. This pattern suggests that Fermi's order-of-magnitude reasoning benefits more from focused computation rather than multi-path exploration. The finding points to a promising research direction: task-aware mindset subsetting, where a minimal effective mindset subset is pre-selected based on problem characteristics, may offer substantial efficiency gains without sacrificing accuracy.



\subsection{Analysis}
\label{sec:analysis}

\begin{figure}[t]
  \centering
  \begin{subfigure}{0.49\textwidth}
    \centering
    \includegraphics[width=\linewidth]{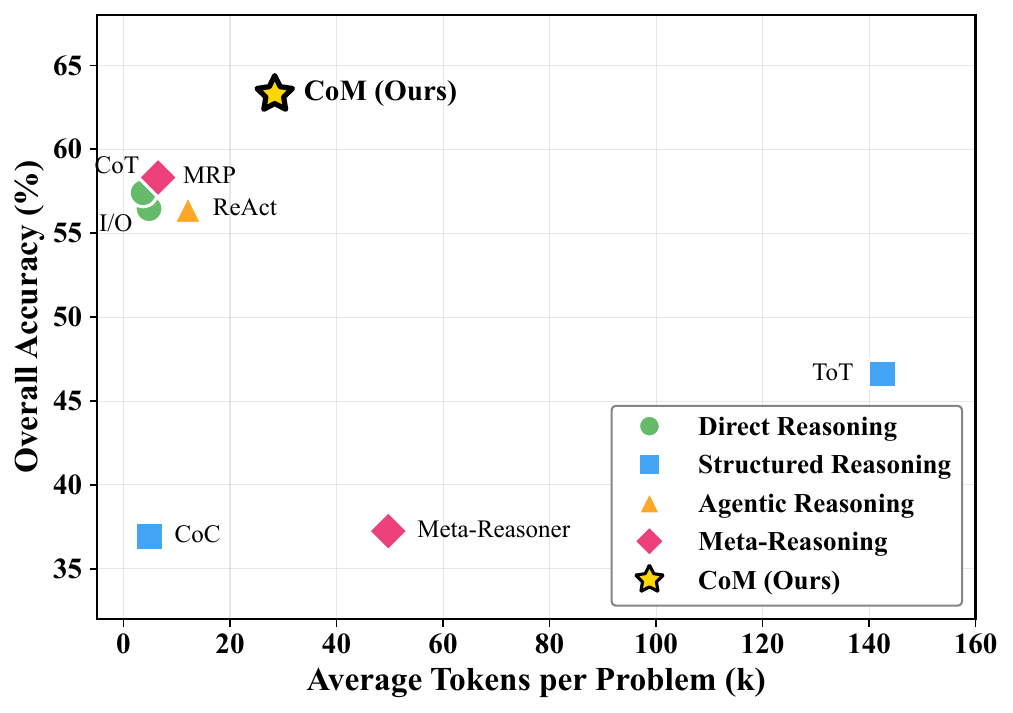}
    \caption{Accuracy-efficiency trade-off across methods on Qwen3-VL-32B-Instruct. Each point represents a method's overall accuracy (\%) vs.\ average token consumption (k). CoM achieves the highest accuracy at moderate cost, dominating the Pareto frontier. Methods in the upper-left region are preferable.}
    \label{fig:efficiency}
  \end{subfigure}\hfill
  \begin{subfigure}{0.49\textwidth}
    \centering
    \includegraphics[width=\linewidth]{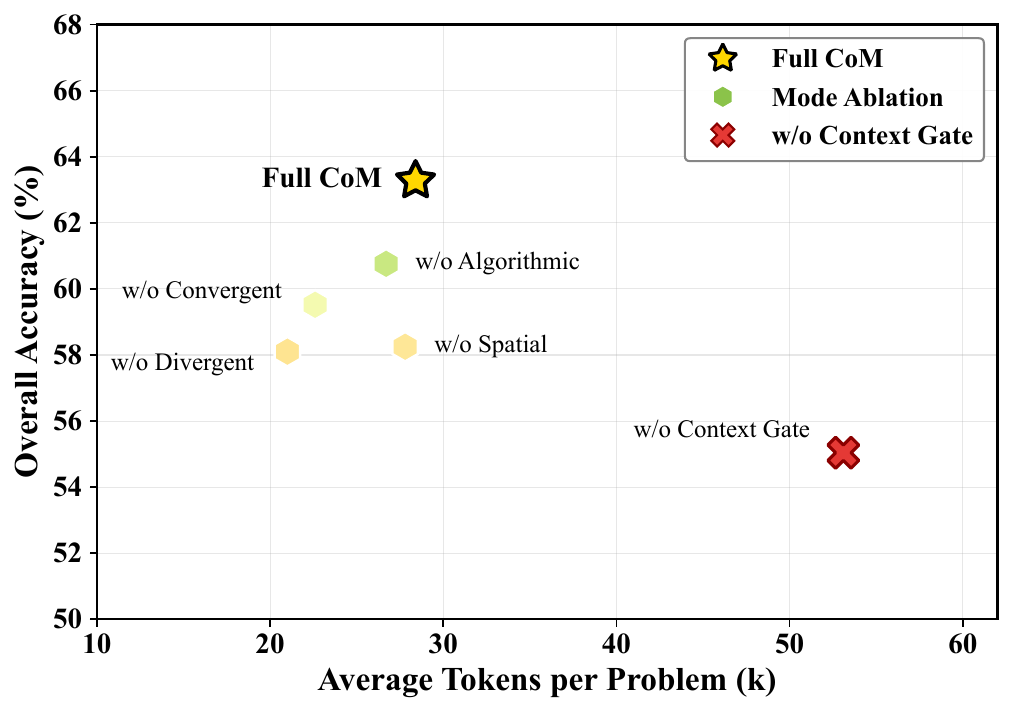}
    \caption{Ablation efficiency trade-off on Qwen3-VL-32B-Instruct. Each point shows a CoM variant's accuracy vs.\ token cost. Removing Context Gate (\texttimes) dramatically increases tokens (+87\%) while degrading accuracy. Removing Divergent mindset offers the best token savings ($-$26\%) with moderate accuracy loss.}
    \label{fig:ablation-efficiency}
  \end{subfigure}
  \caption{Our method achieves state-of-the-art performance while balancing reasoning efficiency.}
  \label{fig:ab}
\end{figure}

\paragraph{Method Efficiency Comparison} We compare CoM against baselines in terms of overall accuracy and token consumption (\cref{fig:efficiency}). Direct methods (Direct I/O, Zero-shot CoT) are most token-efficient but sacrifice substantial accuracy. Tree of Thoughts incurs prohibitive computational cost (142.5k tokens on average) due to exhaustive branch exploration, yet still underperforms CoM in accuracy. Meta-Reasoner also consumes high tokens (49.7k) with relatively low accuracy of 37.25\%. CoM achieves the best accuracy (63.28\%) at moderate cost (28.4k tokens), positioning it on the Pareto frontier of the accuracy-efficiency space.

\paragraph{Ablation Efficiency} Beyond accuracy, we examine the computational cost of each ablation variant (\cref{fig:ablation-efficiency}). Removing the Context Gate increases token consumption by 87\% despite degraded accuracy, as the orchestrator loses its ability to filter irrelevant context. Removing Divergent mode reduces tokens by 26\% with moderate accuracy loss, a viable option for efficiency-critical deployments. The full CoM achieves the best overall accuracy-efficiency trade-off.

\paragraph{Mindset Invocation Patterns} To understand how CoM orchestrates mindsets, we analyze invocation patterns by parsing call sequences recorded during inference. \cref{tab:mode-frequency} reports the percentage of problems invoking each mindset at least once. Overall, 59.7\% of problems invoke two or more distinct mindsets, validating multi-mindset collaboration. Clear task-specific patterns emerge: Fermi estimation relies heavily on Algorithmic (91.2\%) combined with Convergent (78.3\%), reflecting its need for numerical computation and step-by-step analysis. Code generation similarly favors Convergent-Algorithmic combinations, with 60.4\% invoking Algorithmic. Multimodal tasks uniquely leverage Spatial---MathVision at 80.6\% and MAZE at 100\%---demonstrating that CoM adaptively activates visual reasoning for geometric structures.

\begin{table}[t]
\centering
\caption{Mindset invocation frequency (\%) on Qwen3-VL-32B-Instruct. Each cell shows the percentage of problems where the mindset is invoked at least once. ``Multi'' denotes problems invoking two or more distinct mindsets. LCB denotes LiveCodeBench. Overall is the weighted average by problem count.}
\label{tab:mode-frequency}
\small
\begin{tabular}{lcccc|c}
\toprule
Benchmark & Div. & Conv. & Algo. & Spat. & Multi \\
\midrule
AIME25    & 10.0 & 66.7 & 43.3 & 23.3 & 43.3 \\
Fermi     & 70.2 & 78.3 & 91.2 & 13.3 & 88.7 \\
LCB       & 4.4 & 40.1 & 60.4 & 2.2 & 22.5 \\
GPQA      & 23.2 & 74.7 & 39.4 & 14.6 & 51.0 \\
MathV     & 7.6 & 22.4 & 33.6 & 80.6 & 38.8 \\
MAZE      & 0.0 & 4.0 & 39.5 & 100.0 & 40.0 \\
\midrule
\textbf{Overall} & \textbf{34.8} & \textbf{54.5} & \textbf{63.6} & \textbf{33.1} & \textbf{59.7} \\
\bottomrule
\end{tabular}
\end{table}

%% file: section/4.conclusion.tex
\section{Conclusion}

We introduced Chain of Mindset (CoM), a training-free agentic framework that enables step-level adaptive mindset orchestration for LLM reasoning. Unlike existing methods that apply a fixed cognitive strategy throughout problem-solving, CoM dynamically selects among four functionally heterogeneous mindsets, namely Divergent, Convergent, Algorithmic, and Spatial, through a Meta-Agent that responds to the evolving problem state. A bidirectional Context Gate ensures efficient information flow while maintaining focus across mindset transitions. Extensive experiments across six challenging benchmarks demonstrate that CoM achieves state-of-the-art accuracy, outperforming the strongest baselines by 4.96\% and 4.72\% on Qwen3-VL-32B-Instruct and Gemini-2.0-Flash, respectively. Notably, CoM maintains computational efficiency and generalizes consistently across both open-source and closed-source models without requiring additional training. These results suggest that enabling dynamic cognitive switching, which mirrors how humans naturally integrate multiple reasoning modalities within a single problem-solving episode, represents a promising paradigm for building more adaptable reasoning systems.

\newpage

\section*{Impact Statement}
This work advances AI systems that reason more like humans, not by scaling parameters, but by introducing structured cognitive flexibility. CoM shows that orchestrating heterogeneous thinking modes unlocks capabilities beyond single-mode prompting or static strategy selection. Scientifically, our framework provides a testbed for studying interactions among cognitive paradigms, informing both AI and cognitive science. The modular, training-free architecture enables rapid experimentation with new mindsets and policies. We believe \textit{meta-cognitive control}—teaching models to reason about reasoning—is a promising path toward more general intelligence. By making reasoning transparent, CoM enables users to inspect and guide cognitive trajectories, supporting AI that augments human judgment. Regarding safety, explicit reasoning traces enhance auditability, and the structured switching mechanism offers opportunities for targeted safety interventions.

%% file: section/x_supp.tex
\definecolor{mainagentcolor}{RGB}{200, 225, 255}      
\definecolor{mindsetcolor}{RGB}{200, 225, 255}        
\definecolor{gatecolor}{RGB}{255, 240, 220}           
\definecolor{promptbg}{RGB}{248, 248, 248}            

\definecolor{cogdecisioncolor}{RGB}{221, 215, 240}    
\definecolor{algorithmiccolor}{RGB}{214, 238, 250}    
\definecolor{convergentcolor}{RGB}{210, 240, 234}     
\definecolor{divergentcolor}{RGB}{250, 243, 210}      
\definecolor{spatialcolor}{RGB}{225, 240, 210}        
\definecolor{insightcolor}{RGB}{240, 215, 235}        
\definecolor{answercolor}{RGB}{220, 245, 220}         
\definecolor{resultcolor}{RGB}{242, 242, 242}         
\definecolor{inputimgframe}{RGB}{180, 120, 60}        
\definecolor{genimgframe}{RGB}{68, 140, 120}          


\newcommand{\bcogdec}{\colorbox{cogdecisioncolor}{\small\bfseries\texttt{\textless cognitive\_decision\textgreater}}}
\newcommand{\ecogdec}{\colorbox{cogdecisioncolor}{\small\bfseries\texttt{\textless/cognitive\_decision\textgreater}}}

\newcommand{\bcallalgo}{\colorbox{algorithmiccolor}{\small\bfseries\texttt{\textless call\_algorithmic\textgreater}}}
\newcommand{\ecallalgo}{\colorbox{algorithmiccolor}{\small\bfseries\texttt{\textless/call\_algorithmic\textgreater}}}
\newcommand{\bcallconv}{\colorbox{convergentcolor}{\small\bfseries\texttt{\textless call\_convergent\textgreater}}}
\newcommand{\ecallconv}{\colorbox{convergentcolor}{\small\bfseries\texttt{\textless/call\_convergent\textgreater}}}
\newcommand{\bcalldiv}{\colorbox{divergentcolor}{\small\bfseries\texttt{\textless call\_divergent\textgreater}}}
\newcommand{\ecalldiv}{\colorbox{divergentcolor}{\small\bfseries\texttt{\textless/call\_divergent\textgreater}}}
\newcommand{\bcallspat}{\colorbox{spatialcolor}{\small\bfseries\texttt{\textless call\_spatial\textgreater}}}
\newcommand{\ecallspat}{\colorbox{spatialcolor}{\small\bfseries\texttt{\textless/call\_spatial\textgreater}}}

\newcommand{\balgores}{\colorbox{resultcolor}{\small\bfseries\texttt{\textless algorithmic\_result\textgreater}}}
\newcommand{\ealgores}{\colorbox{resultcolor}{\small\bfseries\texttt{\textless/algorithmic\_result\textgreater}}}
\newcommand{\bconvres}{\colorbox{resultcolor}{\small\bfseries\texttt{\textless convergent\_result\textgreater}}}
\newcommand{\econvres}{\colorbox{resultcolor}{\small\bfseries\texttt{\textless/convergent\_result\textgreater}}}
\newcommand{\bdivres}{\colorbox{resultcolor}{\small\bfseries\texttt{\textless divergent\_result\textgreater}}}
\newcommand{\edivres}{\colorbox{resultcolor}{\small\bfseries\texttt{\textless/divergent\_result\textgreater}}}
\newcommand{\bspatres}{\colorbox{resultcolor}{\small\bfseries\texttt{\textless spatial\_result\textgreater}}}
\newcommand{\espatres}{\colorbox{resultcolor}{\small\bfseries\texttt{\textless/spatial\_result\textgreater}}}

\newcommand{\binsight}{\colorbox{insightcolor}{\small\bfseries\texttt{\textless insight\textgreater}}}
\newcommand{\einsight}{\colorbox{insightcolor}{\small\bfseries\texttt{\textless/insight\textgreater}}}

\newcommand{\bans}{\colorbox{answercolor}{\small\bfseries\texttt{\textless Answer\textgreater}}}
\newcommand{\eans}{\colorbox{answercolor}{\small\bfseries\texttt{\textless/Answer\textgreater}}}


\newtcolorbox{casebox}[1][]{
  enhanced,
  colback=white,
  colframe=gray!70,
  boxrule=1pt,
  arc=3pt,
  left=10pt, right=10pt,
  top=8pt, bottom=8pt,
  fonttitle=\bfseries,
  coltitle=black,
  colbacktitle=gray!15,
  title=#1,
  after skip=6pt
}

\newtcolorbox{inputimagebox}[1][IMG\_001]{
  enhanced,
  colback=white,
  colframe=inputimgframe,
  boxrule=1pt,
  arc=2pt,
  left=2pt, right=2pt,
  top=2pt, bottom=2pt,
  title={\footnotesize\ttfamily [#1]},
  coltitle=inputimgframe,
  colbacktitle=white,
  attach boxed title to top right={yshift=-1mm, xshift=-2mm},
  boxed title style={boxrule=0pt, colback=white}
}

\newtcolorbox{genimagebox}[1][GEN\_001]{
  enhanced,
  colback=white,
  colframe=genimgframe,
  boxrule=1pt,
  arc=2pt,
  left=2pt, right=2pt,
  top=2pt, bottom=2pt,
  title={\footnotesize\ttfamily Generated [#1]},
  coltitle=genimgframe,
  colbacktitle=white,
  attach boxed title to top right={yshift=-1mm, xshift=-2mm},
  boxed title style={boxrule=0pt, colback=white}
}


\definecolor{ptframe}{RGB}{74, 144, 217}              
\definecolor{pttext}{RGB}{45, 55, 72}                 
\definecolor{ptaccentblue}{RGB}{49, 130, 206}         
\definecolor{ptaccentgreen}{RGB}{56, 161, 105}        
\definecolor{ptaccentpurple}{RGB}{128, 90, 213}       
\definecolor{ptaccentorange}{RGB}{221, 107, 32}       
\definecolor{ptaccentteal}{RGB}{56, 178, 172}         
\definecolor{ptcodebg}{RGB}{237, 242, 247}            

\newcommand{\psec}[2]{\vspace{0.5em}\noindent\textcolor{#1}{\textbf{#2}}\par\vspace{0.1em}}
\newcommand{\pvar}[1]{\textcolor{ptaccentteal}{\texttt{\{#1\}}}}

\newtcolorbox{instructionbox}[1][]{
  enhanced,
  breakable,
  colback=white,
  colframe=ptframe,
  colbacktitle=ptframe,
  coltitle=white,
  fonttitle=\bfseries\sffamily\small,
  title={\faRobot\hspace{0.5em}#1},
  arc=3pt,
  boxrule=1.2pt,
  left=8pt, right=8pt, top=4pt, bottom=4pt,
  toptitle=2pt, bottomtitle=2pt,
  shadow={1.5pt}{-1.5pt}{0pt}{black!20},
  fontupper=\small\color{pttext},
}

\newtcolorbox{promptcodebox}{
  colback=ptcodebg,
  colframe=ptcodebg,
  arc=2pt,
  boxrule=0pt,
  left=6pt, right=6pt, top=3pt, bottom=3pt,
}

\section{Related Work}

\subsection{Cognitive Behaviors in LLM Reasoning}

Recent research has identified distinct cognitive behaviors in LLM reasoning. Didolkar et al.~\cite{didolkar2024metacognitive} showed that LLMs can identify required skill labels and leverage this self-knowledge to improve performance. Gandhi et al.~\cite{gandhi2025cognitive} identified four key cognitive behaviors—verification, backtracking, subgoal setting, and backward chaining—as critical enablers of self-improvement. Kargupta et al.~\cite{kargupta2025cognitive} introduced a taxonomy of 28 cognitive elements and found that models tend to adopt rigid sequential processing rather than diverse metacognitive monitoring. These works demonstrate that intervening on cognitive behaviors can enhance reasoning, but how to adaptively select the most suitable mindset based on context remains open.

\subsection{Prompt-based Reasoning}

Prompt-based reasoning methods can be categorized into two classes: explicit intermediate step generation and reasoning structure expansion. The former is exemplified by Chain-of-Thought prompting~\cite{wei2022chain}, which improves complex problem-solving by guiding models to generate intermediate steps; Decomposed Prompting~\cite{khot2022decomposed} further decomposes tasks into subtasks delegated to specialized submodules. The latter explores richer reasoning topologies: Program-of-Thoughts~\cite{chen2022program} and Chain-of-Code~\cite{li2023chain} introduce code execution to offload computation; Tree-of-Thoughts~\cite{yao2023tree} and Graph-of-Thoughts~\cite{besta2024graph} employ branching and arbitrary graph structures for multi-path reasoning respectively. These prompting methods enrich reasoning structure and modalities, yet employ a single mindset throughout the task lifecycle—the model remains locked within a predetermined framework. Our method is complementary: while preserving the advantages of prompt-based reasoning, it allows dynamic switching between different mindsets.

\subsection{Meta-Reasoning}

Meta-reasoning—reasoning about how to reason—has emerged as a key paradigm for adaptive strategy selection in LLMs. Existing approaches can be categorized into task-level and step-level methods. Task-level meta-reasoning selects a strategy at problem onset and maintains it throughout: Buffer of Thoughts~\cite{yang2024buffer} retrieves high-level thought templates from a memory library, while MRP~\cite{gao2024mrp} and Sketch-of-Thought~\cite{aytes2025sketch} select the most suitable reasoning paradigm based on problem characteristics. These methods achieve cross-task adaptability but cannot respond to heterogeneous demands of different subtasks within the same problem. Step-level meta-reasoning attempts finer-grained intervention: Meta-Reasoner~\cite{sui2025meta} dynamically schedules execution actions such as backtracking during reasoning; AutoMR~\cite{zhang2025automr} searches for query-aware meta-reasoning skeletons by dynamically expanding DAG structures. Concurrently, Octopus~\cite{guo2025octopus} proposes agentic multimodal reasoning with six-capability orchestration, enabling autonomous capability selection during inference. However, such methods modulate execution parameters or reasoning structures rather than mindsets themselves. Unlike the above work, our method achieves step-level meta-reasoning over functionally heterogeneous mindsets, dynamically determining thinking styles based on subtask context without additional training.

\section{Future Directions}

Our framework instantiates four mindsets representing well-established cognitive primitives. Future work could incorporate additional primitives via our plug-and-play architecture. Currently, all mindsets share the same base model; a natural extension is heterogeneous expert allocation, where each mindset is powered by a specialized model. Additionally, equipping mindsets with tailored tools (e.g., symbolic solvers for Algorithmic, search tools for Convergent) could further enhance capabilities. Finally, optimizing the Meta-Agent's dispatch policy through training could further improve performance.

\clearpage

\section{Baseline Implementation Details}
\label{app:baselines}

For reproducibility, we provide the implementation details of all baseline methods used in our experiments. All methods are evaluated under identical inference settings (temperature, maximum tokens) to ensure fair comparison.

\begin{itemize}
    \item \textbf{Direct I/O.} Direct I/O queries the model without any reasoning guidance or system prompt. The model receives only the question and format instruction (if provided by the dataset), representing the minimal baseline for comparison. Prompt template: \texttt{"Question: \{question\} Answer:"}
    
    \item \textbf{Zero-shot CoT.} Zero-shot CoT~\citep{kojima2022large} elicits chain-of-thought reasoning by appending the canonical trigger phrase to the question. Following the original paper, we use: \texttt{"Question: \{question\} Let's think step by step."}
    
    \item \textbf{Tree of Thoughts.} We implement Tree of Thoughts~\citep{yao2023tree} with BFS/Beam Search strategy. The model decomposes problems into sub-questions, generates $k=3$ candidate thoughts per step, evaluates each candidate's usefulness and correctness via self-evaluation, and selects the best branch to expand. Maximum reasoning depth is set to 10 steps.
    
    \item \textbf{Chain of Code.} We implement Chain of Code~\citep{li2023chain} following the original paper. The model generates Python code to solve problems and simulates code execution. If actual execution fails (timeout or exception), the model's predicted output is used as fallback. Execution timeout is set to 10 seconds.
    
    \item \textbf{ReAct.} We implement the ReAct framework~\citep{yao2022react} with the standard Thought-Action-Observation loop. For fair comparison, we equip ReAct with the same tool set as CoM: (1) \textbf{PythonSandbox}: for code execution and numerical computation (timeout: 30 seconds); (2) \textbf{ImageGeneration}: for visualization using the same image generation API as CoM's Spatial mode. Maximum interaction turns is set to 10.
    
    \item \textbf{MRP.} MRP~\citep{gao2024mrp} does not have open-source code, but provides prompts in the original paper. We follow the paper to implement MRP. At reasoning onset, the model analyzes problem characteristics through meta-reasoning, rates the suitability of each method (Chain-of-Thoughts, Tree-of-Thoughts, Analogical Prompting, Self-Refine, Step-Back Prompting, Solo Performance Prompting, SimTom) on a 1--7 scale, and selects the highest-scoring method for execution.
    
    \item \textbf{Meta-Reasoner.} Meta-Reasoner~\citep{sui2025meta} does not have open-source code, but provides prompts, pseudo code, and detailed description in the original paper. We follow the paper to implement Meta-Reasoner. It uses contextual multi-armed bandits to dynamically select control actions (continue, backtrack, restart, etc.) during reasoning, with exploration rate $\epsilon=0.1$.
\end{itemize}

\clearpage

\section{Chain of Mindsets Prompt Templates}
\label{app:com_prompts}

We provide the complete prompt templates used in Chain of Mindsets (CoM). Our framework consists of a Main Agent (Meta-Cognitive Orchestrator), four specialized Mindset Experts, and Context Gates for information filtering.

\subsection{Meta-Agent}

\begin{instructionbox}[Meta-Agent System Prompt]

\psec{ptaccentpurple}{\faCogs\hspace{0.3em}Role Definition}

You are a \textbf{Meta-Cognitive Orchestrator}. You decide \textbf{HOW} to think, not WHAT to think. Delegate all reasoning to cognitive modules.

\psec{ptaccentblue}{\faCubes\hspace{0.3em}Cognitive Modules}

\begin{promptcodebox}
\begin{itemize}[leftmargin=1.2em, itemsep=1pt, label=\textcolor{ptaccentblue}{\faCaretRight}]
  \item \textbf{Algorithmic} \texttt{\textless call\_algorithmic\textgreater...\textless/call\_algorithmic\textgreater}: Precise calculations and verifications
  \item \textbf{Spatial} \texttt{\textless call\_spatial\textgreater...\textless/call\_spatial\textgreater}: Structures and spatial relationships
  \item \textbf{Divergent} \texttt{\textless call\_divergent\textgreater...\textless/call\_divergent\textgreater}: Multiple solution paths in parallel
  \item \textbf{Convergent} \texttt{\textless call\_convergent\textgreater...\textless/call\_convergent\textgreater}: Deep logical analysis on a sub-question
\end{itemize}
\vspace{-0.3em}
Reference images as \texttt{[IMG\_001]}, \texttt{[GEN\_001]} when relevant.
\end{promptcodebox}

\psec{ptaccentgreen}{\faListOl\hspace{0.3em}Protocol}

Be concise. Never reason in \texttt{\textless cognitive\_decision\textgreater} --- only plan which mindsets to use.
Be concise. Never reason in \texttt{\textless call\_xxx\textgreater} --- only call.
Execute mindsets in planned order. Monitor history; revise unexecuted plan anytime via \texttt{\textless cognitive\_decision\textgreater}.

\begin{itemize}[leftmargin=1.2em, itemsep=1pt, label=\textcolor{ptaccentgreen}{\faCaretRight}]
  \item \texttt{\textless cognitive\_decision\textgreater...\textless/cognitive\_decision\textgreater} --- Identify problem type + plan mindsets. No solving.
  \item \texttt{\textless call\_xxx\textgreater...\textless/call\_xxx\textgreater} $\rightarrow$ \texttt{\textless xxx\_result\textgreater...\textless/xxx\_result\textgreater} $\rightarrow$ \texttt{\textless insight\textgreater...\textless/insight\textgreater} --- Call, receive, internalize briefly.
  \item \texttt{\textless Answer\textgreater...\textless/Answer\textgreater} --- Final response.
\end{itemize}

\psec{ptaccentorange}{\faLightbulb\hspace{0.3em}Example}

\begin{promptcodebox}
Q: ``A bee flies between two trains 300km apart (speeds 60, 90 km/h) at 120 km/h until they meet. Distance?''

\smallskip
\textless cognitive\_decision\textgreater\
Pursuit problem with oscillation. Convergent $\rightarrow$ Algorithmic $\rightarrow$ Convergent.
\textless/cognitive\_decision\textgreater

\textless call\_convergent\textgreater Model the bee's path.\textless/call\_convergent\textgreater

\textless convergent\_result\textgreater Infinite series: sum segments as bee bounces.\textless/convergent\_result\textgreater

\textless insight\textgreater Series approach identified.\textless/insight\textgreater

\textless call\_algorithmic\textgreater Compute first segments.\textless/call\_algorithmic\textgreater

\textless algorithmic\_result\textgreater Seg1: 144km, Seg2: 57.6km... continues.\textless/algorithmic\_result\textgreater

\textless insight\textgreater Tedious. Simpler way?\textless/insight\textgreater

\smallskip
\textless cognitive\_decision\textgreater\
Method too complex. Divergent $\rightarrow$ Algorithmic.
\textless/cognitive\_decision\textgreater

\textless call\_divergent\textgreater Alternative approaches?\textless/call\_divergent\textgreater

\textless divergent\_result\textgreater A: Sum series. B: Total flight time = meeting time. C: Relative velocity.\textless/divergent\_result\textgreater

\textless insight\textgreater B: just compute meeting time.\textless/insight\textgreater

\textless call\_algorithmic\textgreater 300/(60+90) = 2h. 120 $\times$ 2 = ?\textless/call\_algorithmic\textgreater

\textless algorithmic\_result\textgreater 240 km.\textless/algorithmic\_result\textgreater

\textless insight\textgreater Done.\textless/insight\textgreater

\textless Answer\textgreater 240 km\textless/Answer\textgreater
\end{promptcodebox}

Begin.
\end{instructionbox}

\subsection{Mindset Experts}

\paragraph{Algorithmic Mindset.}
The Algorithmic Mindset handles precise calculations and code-based verifications. It generates executable Python code and supports self-correction on errors.

\begin{instructionbox}[Algorithmic Mindset: Code Generation Prompt]

\psec{ptaccentpurple}{\faBullseye\hspace{0.3em}Task}

You are verifying through computation.

\begin{promptcodebox}
\pvar{instruction}
\end{promptcodebox}

\psec{ptaccentgreen}{\faCode\hspace{0.3em}Instructions}

Write executable Python code that solves the task precisely. Print the result to stdout.

\end{instructionbox}

\begin{instructionbox}[Algorithmic Mindset: Code Fix Prompt]

The previous code failed.

\psec{ptaccentblue}{\faCode\hspace{0.3em}Code}

\begin{promptcodebox}
\pvar{code}
\end{promptcodebox}

\psec{ptaccentorange}{\faExclamationTriangle\hspace{0.3em}Error}

\begin{promptcodebox}
\pvar{error}
\end{promptcodebox}

\psec{ptaccentgreen}{\faWrench\hspace{0.3em}Action}

Fix the code. Preserve the original intent.

\end{instructionbox}

\paragraph{Convergent Mindset.}
The Convergent Mindset performs deep logical analysis on focused questions, emphasizing rigorous reasoning grounded in established facts.

\begin{instructionbox}[Convergent Mindset System Prompt]

\psec{ptaccentpurple}{\faBullseye\hspace{0.3em}Role}

You are reasoning deeply.

\psec{ptaccentgreen}{\faListOl\hspace{0.3em}Guidelines}

\begin{itemize}[leftmargin=1.2em, itemsep=1pt, label=\textcolor{ptaccentgreen}{\faCaretRight}]
  \item Focus all attention on the given question.
  \item Ground each step in established facts.
  \item If information is insufficient, state what is missing.
  \item Reach a clear conclusion.
\end{itemize}

\end{instructionbox}

\paragraph{Divergent Mindset.}
The Divergent Mindset explores multiple solution paths in parallel. It first generates diverse approaches, then performs deep-dive exploration on each branch.

\begin{instructionbox}[Divergent Mindset: Branch Generation Prompt]

\psec{ptaccentpurple}{\faBullseye\hspace{0.3em}Task}

You are exploring possibilities.

\begin{promptcodebox}
\pvar{instruction}
\end{promptcodebox}

Generate 2--5 genuinely different approaches. Each approach should differ in method, not just phrasing.

\psec{ptaccentorange}{\faFileExport\hspace{0.3em}Output Format}

\begin{promptcodebox}
\texttt{\textless branch id="A"\textgreater} \\
\quad [Approach name] \\
\quad Method: [How it works] \\
\quad When applicable: [Conditions] \\
\texttt{\textless/branch\textgreater}

\texttt{\textless branch id="B"\textgreater} \\
\quad ... \\
\texttt{\textless/branch\textgreater}
\end{promptcodebox}

\end{instructionbox}

\begin{instructionbox}[Divergent Mindset: Branch Exploration Prompt]

\psec{ptaccentpurple}{\faBullseye\hspace{0.3em}Task}

You are exploring one approach in depth.

\psec{ptaccentblue}{\faDatabase\hspace{0.3em}Context}

\begin{promptcodebox}
\textbf{Problem:} \pvar{instruction} \\
\textbf{Approach:} \pvar{branch\_description}
\end{promptcodebox}

\psec{ptaccentgreen}{\faListOl\hspace{0.3em}Examination Steps}

\begin{enumerate}[leftmargin=1.2em, itemsep=1pt]
  \item What assumptions does it make? Are they satisfied?
  \item How would it work step by step?
  \item What are the limitations?
  \item Is it viable for this problem?
\end{enumerate}

Be honest about limitations.

\end{instructionbox}








\paragraph{Spatial Mindset.}
The Spatial Mindset handles visual-spatial thinking, transforming abstract descriptions into visual representations. Unlike other mindsets that use explicit prompts, the Spatial Mindset directly routes the processed context to an image generation model. The Main Agent's call instruction (e.g., ``Visualize the geometric relationship'') is first processed by the Input Gate, which extracts relevant context and decides which reference images to inject. The combined context and instruction are then sent to an image generation API (we use Nano Banana Pro with native image generation capabilities). The generated image is saved to the session workspace, and the Output Gate extracts the image path along with any accompanying notes for the Main Agent.

\textbf{Supported Modes:}
\begin{itemize}
    \item \textbf{Text $\rightarrow$ Image}: Pure text description generates visualization
    \item \textbf{Image + Text $\rightarrow$ Image}: Reference images [IMG\_XXX] are injected for editing/redrawing
    \item \textbf{Text/(Image + Text)$\rightarrow$ Code $\rightarrow$ Image}: If the API returns matplotlib code instead of an image, the code is executed in a sandbox to generate the figure
\end{itemize}

\subsection{Context Gates}

The Context Gate implements a cognitive gating mechanism inspired by the PBWM (Prefrontal Cortex Basal Ganglia Working Memory) model from cognitive neuroscience~\cite{o2006making}. The same \texttt{call} serves as the anchor point for both directions, ensuring information relevance throughout the cognitive loop.

\paragraph{Input Gate.}
The Input Gate filters information from Main Agent history to extract only what the specialized Mindset needs, and decides which images to inject based on semantic relevance.

\begin{instructionbox}[Input Gate Prompt]

\psec{ptaccentpurple}{\faFilter\hspace{0.3em}Role}

You are the attentional filter of a cognitive agent. Extract what the specialized thinking needs to execute the instruction.

\psec{ptaccentblue}{\faDatabase\hspace{0.3em}Input Context}

\begin{promptcodebox}
\textbf{Instruction:} \pvar{call} \\
\textbf{History:} \pvar{source\_history} \\
\textbf{Available Images:} \pvar{available\_images\_description} \\
\textbf{Target:} \pvar{target\_description}
\end{promptcodebox}

\psec{ptaccentgreen}{\faListOl\hspace{0.3em}Extraction Rules}

\begin{itemize}[leftmargin=1.2em, itemsep=1pt, label=\textcolor{ptaccentgreen}{\faCaretRight}]
  \item \textbf{Keep verbatim}: numbers, data, coordinates, prior results, text being analyzed.
  \item \textbf{Summarize}: reasoning chains $\rightarrow$ conclusions only.
  \item \textbf{Omit}: the original user question (the thinking sees only its sub-task).
\end{itemize}

\psec{ptaccentblue}{\faImage\hspace{0.3em}Image Decision}

\begin{itemize}[leftmargin=1.2em, itemsep=1pt, label=\textcolor{ptaccentblue}{\faCaretRight}]
  \item Explicit \texttt{[IMG\_XXX]} in instruction $\rightarrow$ inject those
  \item ``the figure/image'' without marker $\rightarrow$ inject most relevant
  \item Purely textual task $\rightarrow$ inject nothing
\end{itemize}

\psec{ptaccentorange}{\faFileExport\hspace{0.3em}Output Format (JSON only)}

\begin{promptcodebox}
\texttt{\{``context\_text'': ``extracted context or empty string'', ``inject\_images'': [``IMG\_001''] or []\}}
\end{promptcodebox}

\end{instructionbox}

\begin{instructionbox}[Input Gate: Target Descriptions]

\begin{itemize}[leftmargin=1.2em, itemsep=2pt, label=\textcolor{ptaccentblue}{\faCaretRight}]
  \item \textbf{Algorithmic Mindset} --- executes precise calculations and code-based verifications
  \item \textbf{Convergent Mindset} --- performs deep logical analysis on focused questions
  \item \textbf{Divergent Mindset} --- explores multiple approaches and alternatives in parallel
  \item \textbf{Spatial Mindset} --- creates and analyzes visual-spatial representations
\end{itemize}

\end{instructionbox}

\paragraph{Output Gate.}
The Output Gate extracts results from Mindset execution that advance the main reasoning, filtering out derivation steps and failed attempts.

\begin{instructionbox}[Output Gate Prompt]

\psec{ptaccentpurple}{\faFilter\hspace{0.3em}Role}

You are the attentional filter of a cognitive agent. Extract the results that advance the main reasoning.

\psec{ptaccentblue}{\faDatabase\hspace{0.3em}Input Context}

\begin{promptcodebox}
\textbf{Instruction:} \pvar{call} \\
\textbf{Execution Record:} \pvar{mindset\_output} \\
\textbf{New Artifacts:} \pvar{new\_artifacts\_description}
\end{promptcodebox}

\psec{ptaccentgreen}{\faListOl\hspace{0.3em}Extraction Rules}

\begin{itemize}[leftmargin=1.2em, itemsep=1pt, label=\textcolor{ptaccentgreen}{\faCaretRight}]
  \item \textbf{Keep}: computed values, discovered patterns, conclusions, generated image paths.
  \item \textbf{Omit}: derivation steps, failed attempts.
\end{itemize}

\textbf{Extracted Results}

\end{instructionbox}

\clearpage

\section{Case Studies}
\label{app:case_studies}

We present two additional representative case studies demonstrating Chain of Mindsets (CoM) across different problem types: (1) mathematical reasoning with dynamic re-planning, and (2) multimodal geometry with visual input. Special tokens are highlighted (e.g., \bcogdec, \bcallconv) to show the meta-cognitive control flow. An additional Fermi estimation example demonstrating the Spatial Mindset's image generation capability is provided in Section~\ref{sec:case_example}.

\subsection{Case Study 1: Mathematical Reasoning with Dynamic Re-planning (AIME)}

\begin{casebox}[AIME 2025: Base Divisibility]

\textbf{Question:} Find the sum of all integer bases $b > 9$ for which $17_b$ is a divisor of $97_b$.

\textbf{Answer:} \textbf{70}

\tcbline

\bcogdec Base conversion followed by divisibility check. Plan: Convergent $\rightarrow$ Algorithmic. \ecogdec

\bcallconv Express $17_b$ and $97_b$ in base 10. \ecallconv

\bconvres $17_b = b + 7$, $97_b = 9b + 7$. Condition: $(b+7) \mid (9b+7)$. \econvres

\binsight The divisibility condition $(b+7) \mid (9b+7)$ can be simplified algebraically before testing values. \einsight

\bcogdec Remaining plan was Algorithmic (test bases). But the condition can be reduced first. Revise to: Convergent $\rightarrow$ Algorithmic. \ecogdec

\bcallconv Simplify $(b+7) \mid (9b+7)$. \ecallconv

\bconvres $9b + 7 = 9(b+7) - 56$, so $(b+7) \mid 56$. \econvres

\binsight Now find divisors of 56 greater than 16 (since $b = d - 7 > 9$ requires $d > 16$). \einsight

\bcallalgo Divisors of 56: \{1,2,4,7,8,14,28,56\}. Filter $d > 16$: \{28, 56\}. Compute $b = d - 7$. \ecallalgo

\balgores $b = 21$ and $b = 49$. Sum: $21 + 49 = 70$. \ealgores

\binsight Final answer: 70. \einsight

\bans \textbf{70} \eans

\end{casebox}

This example illustrates CoM's core capability: state-dependent cognitive switching. Unlike static meta-reasoning that commits to a fixed strategy, CoM monitors intermediate results and dynamically revises its plan when a more efficient path emerges. Here, the first Convergent call formulates the divisibility condition, and the insight recognizes that algebraic simplification should precede enumeration. The second Convergent call then reduces $(b+7) \mid (9b+7)$ to $(b+7) \mid 56$, enabling efficient computation via Algorithmic mindset.

\subsection{Case Study 2: Multimodal Geometry with Visual Input (MathVision)}

\begin{casebox}[MathVision \#1167: Zig-Zag Angles]

\begin{minipage}[c]{0.55\textwidth}
\textbf{Question:} Valentin draws a zig-zag line inside a rectangle using angles $10°$, $14°$, $33°$ and $26°$. How big is angle $\varphi$?

\textbf{Options:} A. $11°$ \quad B. $12°$ \quad C. $16°$ \quad D. $17°$ \quad E. $33°$

\textbf{Answer:} \textbf{A}
\end{minipage}
\hfill
\begin{minipage}[c]{0.42\textwidth}
\vspace{0pt}
\begin{inputimagebox}[IMG\_001]
\centering
\includegraphics[width=0.88\linewidth]{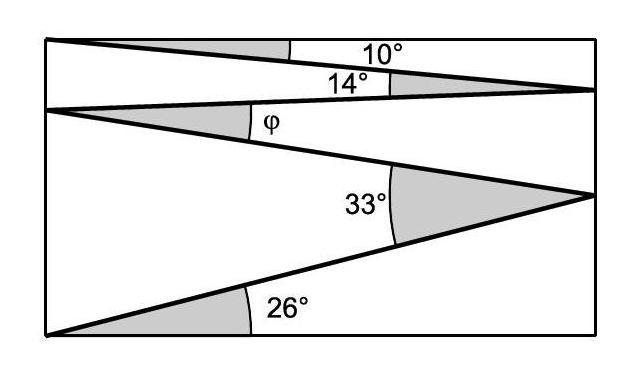}
\end{inputimagebox}
\end{minipage}

\tcbline

\bcogdec Geometry problem with angles in a zig-zag path. Plan: Convergent $\rightarrow$ Algorithmic. \ecogdec

\bcallconv What geometric relationship governs the zig-zag angles? \ecallconv

\bconvres Net direction change = $33° - 10° = 23°$. Net turning = $\varphi - 21°$. Equation: $\varphi - 21° = 23°$ $\Rightarrow$ $\varphi = 44°$. \econvres

\binsight $44°$ not in options. The model is flawed. \einsight

\bcogdec Initial approach failed. Need alternative geometric principle. Revise to: Divergent $\rightarrow$ Algorithmic. \ecogdec

\bcalldiv Alternative geometric principles for zig-zag in rectangle? \ecalldiv

\bdivres \textbf{A}: Cumulative direction change --- already failed. \textbf{B}: Zig-zag theorem: sum of left-pointing angles = sum of right-pointing angles. \textbf{C}: Exterior angle sum --- not applicable. \edivres

\binsight Branch B: Zig-zag theorem looks promising. \einsight

\bcallalgo Apply zig-zag theorem. Right angles: $10° + \varphi + 26°$. Left angles: $14° + 33°$. Solve. \ecallalgo

\balgores $36° + \varphi = 47°$ $\Rightarrow$ $\varphi = 11°$. \ealgores

\binsight $11°$ matches option A. Done. \einsight

\bans \textbf{A} \eans

\end{casebox}

This example demonstrates CoM's error recovery through mindset switching. When the initial Convergent approach yields an answer ($44°$) absent from the options, the insight mechanism detects the inconsistency and triggers re-planning. The subsequent Divergent call generates alternative geometric principles, among which the zig-zag theorem proves viable. The Algorithmic mindset then executes the correct calculation, illustrating how CoM leverages mindset diversity to escape reasoning dead-ends.